\documentclass[review, journal]{IEEEtran}

\usepackage{amsfonts}
\usepackage{amsmath}
\usepackage{amsmath,epsfig,amsmath}
\usepackage{amsfonts,amssymb,amsthm}
\usepackage{epsfig}
\usepackage{float}
\usepackage{graphicx}
\usepackage{color,hyperref}
\usepackage{multirow}
\usepackage{tipa}
\usepackage{picins}
\usepackage{color}
\usepackage{booktabs}
\usepackage[table]{xcolor}
\usepackage{threeparttable}

\def\eg{\emph{e.g.}} 
\def\ie{\emph{i.e.}}

\begin{document}

\title{Cognitive Accident Prediction in Driving Scenes: \\A Multimodality Benchmark}

\author{Jianwu Fang, Lei-Lei Li, Kuan Yang, Zhedong Zheng, Jianru Xue, and Tat-Seng Chua
\thanks{

J. Fang, L. Li, and K. Yang are with the College of Transportation Engineering, Chang'an University, Xi'an, China, and J. Fang is also a visiting scholar in NExT++ Research Centre of School of Computing, National University of Singapore, Singapore.
        {(fangjianwu@chd.edu.cn)}.}%
        
   \thanks{Z. Zheng and T-S. Chua are with the Sea-NExT Joint Research Centre of School of Computing, National University of Singapore, Singapore.
        {(zdzheng,dcscts@nus.edu.sg)}.}
\thanks{J. Xue are with the Institute of Artificial Intelligence and Robotics, Xi'an Jiaotong University, Xi'an, China
        {(jrxue@mail.xjtu.edu.cn)}.}%
     
}

\markboth{}%
{Shell \MakeLowercase{\textit{et al.}}: Bare Demo of IEEEtran.cls for Computer Society Journals}
%



\IEEEtitleabstractindextext{%
\begin{abstract}
Traffic accident prediction in driving videos aims to provide an early warning of accident occurrence, and supports the decision-making of safe driving systems. Previous works usually concentrate on the spatial-temporal correlation of object-level context, while they do not fit the inherent long-tailed data distribution well and are vulnerable to severe environmental change. In this work, we propose a Cognitive Accident Prediction (CAP) method that explicitly leverages human-inspired cognition of text description on the visual observation and the driver attention to facilitate model training. 
In particular, the text description provides a dense semantic description guidance for the primary context of the traffic scene, while the driver attention provides traction to focus on the critical region closely correlating with safe driving. CAP is formulated by an attentive text-to-vision shift fusion module, an attentive scene context transfer module, and the driver attention-guided accident prediction module. We leverage the attention mechanism in these modules to explore the core semantic cues for accident prediction. In order to train CAP, we extend an existing self-collected DADA-2000 dataset (with annotated driver attention for each frame) with further factual text descriptions for the visual observations before the accidents. Besides, we construct a new large-scale benchmark consisting of 11,727 in-the-wild accident videos with over 2.19 million frames (named as \textbf{CAP-DATA}) together with labeled \emph{fact-effect-reason-introspection} description and temporal accident frame label. Based on extensive experiments, the superiority of CAP is validated compared with state-of-the-art approaches.
\end{abstract}

\begin{IEEEkeywords}
Traffic accident prediction, Safe driving, Cognitive mechanism, Attentive network, Transformer
\vspace{4em}
\end{IEEEkeywords}}

\maketitle

\IEEEdisplaynontitleabstractindextext

%
\IEEEpeerreviewmaketitle

\begin{figure}[!t]
  \centering
 \includegraphics[width=0.96\hsize]{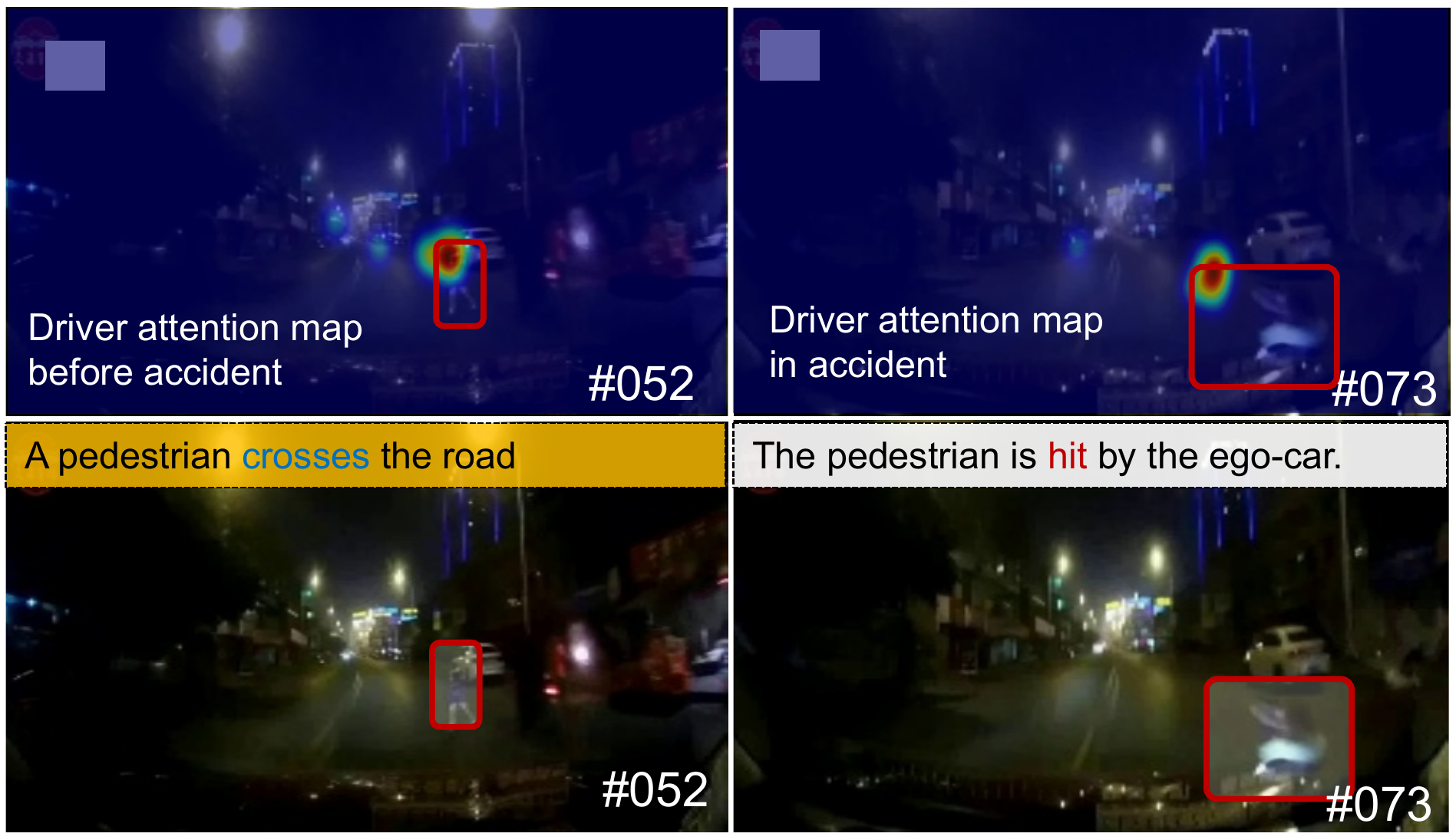}
  \caption{A typical pedestrian accident in DADA-2000 benchmark \cite{DBLP:journals/tits/FangYQXY22}. The pedestrian (marked by the \textcolor{red}{red} box) in the accident is hard to be detected because of the low light condition and irregular shape. Nevertheless, we observe that the text description and the driver attention usually can serve as an accurate prompt.}
  \label{fig1}
\end{figure}

\IEEEraisesectionheading{
\section{Introduction}
\label{section1}}
\IEEEPARstart{R}{oad} accidents account for a significant portion of the annual number of fatalities in the world, especially among young people \cite{trafficdeath}.
In order to avoid such accidents, traffic accident prediction aims to forecast the future traffic accident in advance, and make timely decisions for successful collision avoidance. Facing this goal, current autonomous driving and assisted driving systems have devoted more and more efforts to predict the status of road participants in surroundings, such as the trajectory prediction \cite{DBLP:journals/corr/abs-2109-12764,su2022trajectory,DBLP:conf/iros/MerschHZSR21,DBLP:journals/corr/abs-2211-00848,zheng2020vehiclenet}, action prediction \cite{DBLP:journals/ijcv/WuWHLL21,DBLP:conf/ijcai/YaoAJV021,DBLP:journals/tcsv/ChenLSZ21}, and behavioral intention prediction \cite{DBLP:journals/corr/abs-2211-00385, DBLP:journals/corr/abs-2209-10767}, \emph{etc}. 
However, one inherent challenge still remains: traffic accidents often occur in a very short time window and at extraordinary regions. Learning such clues demands mining the sparse spatial-temporal semantics of accidents and being aware of the sudden inconsistent shifts from the normal condition to an accident situation, as shown in Fig. \ref{fig1}.
In addition, the varying weather and light conditions, highly-imbalanced accident types, and numerous occasions increase the prediction difficulty.

Recently, a number of traffic accident prediction models in dashcam videos have emerged \cite{DBLP:conf/accv/ChanCXS16,DBLP:conf/cvpr/SuzukiKAS18,karim2022attention} to predict the occurrence label of the beginning time of an accident with the preference of larger Time-to-Accident (TTA). Scrutinizing these models, the common paradigms concentrate on the consistency or correlation of the object tracklets \cite{DBLP:conf/cvpr/SuzukiKAS18,karim2022attention} or visual context \cite{DBLP:journals/tits/KarimLQY22,bao2020uncertainty,chen2022pipa} in deep learning. The spatial interaction model is commonly adopted to explore the visual context, such as the spatial-temporal attention networks \cite{DBLP:journals/tits/KarimLQY22,DBLP:conf/accv/ChanCXS16} that take the primary pipeline for video inference for accident situations. However, it is common sense that accidents are more likely to occur in the severe environment conditions, during which the target objects are difficult to be detected or tracked, and the feasibility of these mainstream methods is limited. Most recently, Bao and Kong \cite{DBLP:conf/iccv/Bao0K21} propose a deep reinforced accident anticipation model with fixation guidance, while they pre-train the driver attention map offline, and do not explore the text description that is promising for cognition-inspired video reasoning problems \cite{DBLP:conf/mm/LiWZZZZMPW22,DBLP:conf/cvpr/0004WXJC22,DBLP:conf/mm/00040XC22,DBLP:conf/aaai/XiaoYL0JC22}.

In this work, we explore human cognition clues to assist accident prediction in driving videos. In particular, we explore the text description before the accident, as well as the driver attention in driving situations to \textbf{boost the accuracy and the explainability} of the accident prediction model, which has the apparent semantic guidance for the object to be involved in the accident and helps to find the crashing object efficiently. From Fig. \ref{fig1}, without the textual description and the driver attention guidance, the pedestrian is hard to be found. Actually, as investigated by the gaze data-based exploratory study \cite{DBLP:journals/corr/abs-2108-01599}, AI models and humans can collaboratively improve road safety in conditional automation but effective computation models are under-explored.

In order to fulfill this purpose, we extend the self-collected DADA-2000 benchmark \cite{DBLP:journals/tits/FangYQXY22} (with driver attention map for each frame) with text description for the video frames before the accident, as well as the label at the beginning time of the collision. 
In addition, we also partition each video sequence into different accident categories. The extended factual description before the accident and the driver attention map is adopted to learn a Cognitive Accident Prediction (CAP) model. 
In addition to DADA-2000, we also construct a \textbf{new large-scale accident benchmark} (named CAP-DATA) for performance evaluation in this work. It contains 11,727 in-the-wild dashcam videos with \textbf{2.19 million} frames. 
We annotate the \emph{fact-effect-reason-introspection} description for each accident video, which consists of the factual description before the accident, the categorical accident description, the accident reason description, and the preventive advice description. To our best knowledge, this is the largest accident prediction benchmark in driving scenarios. 

Based on the new benchmarks, we introduce a new vision-text fusion model, called Cognitive Accident Prediction (CAP). In particular, the proposed CAP model consists of three core modules: attentive text-to-vision shift fusion module, attentive semantic context transfer module, and driver attention guided accident prediction module. It is worth noting that we learn the coherence of text-vision semantics by cascade attentive networks, where each attentive module fulfills the core semantic learning for accident prediction, and contributes human-inspired cognitive modeling. Attentive text-to-vision shift fusion (Sec.~\ref{at2vsf}) is modeled by inferring the coherently semantic relation of text and video for accident prediction. Besides, the text-to-video shift strategy aims to leverage the semantic knowledge in the text to video, so as to adapt to the practical testing phase only with the video data. The attentive semantic context transfer module (Sec.~\ref{asct}) encodes the scene context that is modeled by the Graph Neural Network (GNN) and Gated Recurrent Unit (GRU), which imitates the ability of human beings for historical and contextual memory learning. Finally, the driver attention guided accident prediction module (Sec.~\ref{dagap}) forecasts the beginning time of the accident and reconstructs the driver attention map simultaneously. Extensive experiments show that the proposed method can obtain a \textbf{more accurate} prediction with a \textbf{larger Time-to-Accident (TTA)} than other state-of-the-art methods \footnote{The code, CAP-DATA, and all results will be released in \url{https://github.com/JWFanggit/LOTVS-CAP}}. 

The remainder of this work is organized as follows. Sec.~\ref{relatework} briefly reviews the related works. CAP-DATA is presented in Sec.~\ref{ben}, and the proposed method in Sec.~\ref{method}. We evaluate the proposed method and analyze the results in Sec.~\ref{experiments} followed by the conclusion in Sec.~\ref{con}.
\begin{table*}[!t]\footnotesize
  \centering
  \caption{Characteristics of the real-world accident datasets. TA.: temporal annotation of the accident, Surv.:surveillance view.}
  \renewcommand{\arraystretch}{1.2}
 \setlength{\tabcolsep}{1.8mm}{
\begin{tabular}{c|ccccccccc}
\toprule[0.8pt]
Datasets &view View& \#clips &\#frames &attention&tracklet&TA.& \#ave-frames&trimmed&causal description\\
\hline
CADP \cite{shah2018accident} $_\emph{AVSS2018}$&Surv.& 1416& 51.8k&  & & \checkmark& 366&no&\\
SUTD-TrafficQA \cite{DBLP:conf/cvpr/XuHL21} $_\emph{CVPR2021}$&Surv.& 10,080& &  &  & \checkmark& &no&\checkmark\\
 \hline
DAD \cite{DBLP:conf/accv/ChanCXS16} $_\emph{ACCV2016}$&Dashcam& 1750& 175k&  &  \checkmark & \checkmark& 100&yes&\\
A3D  \cite{DBLP:conf/iros/YaoXWCA19} $_\emph{ICRA2019}$ &Dashcam& 3757& 208k& & & \checkmark &  139&no&\\
CCD \cite{bao2020uncertainty} $_\emph{ACMMM2020}$ &Dashcam& 1381&75k&   & \checkmark & \checkmark&  50&yes&\\
Eyecar \cite{DBLP:conf/iccv/BaeePK0OB21} $_\emph{ICCV2021}$ &Dashcam& 20&315k&\checkmark & & & 15.7k&no& \\
DADA-2000 \cite{DBLP:journals/tits/FangYQXY22} $_\emph{IEEE-TITS2022}$ &Dashcam& 2000&658k&\checkmark  &  & \checkmark &  329&no&\checkmark\\
DoTA \cite{yao2022dota} $_\emph{IEEE-TPAMI2022}$ &Dashcam& 5586&732k&   &  & \checkmark&  156&no&\\
\textbf{CAP-DATA} &Dashcam& \textbf{11,727}&\textbf{2.19 millon}& partial  &  & \checkmark&  187 &no&\checkmark\\
  \hline
  \end{tabular}}
  \label{tab1}
  \end{table*}
  
\section{Precedent Work}
\label{relatework}
This work briefly describes the precedent works of accident anticipation, explainable AI (XAI) in accident prediction, and accident prediction datasets in driving scenarios. 

\subsection{Accident Anticipation in Driving Scenarios}
Accident Anticipation can be formulated as a probability prediction of a future accident, where the problem is the same as future or early action prediction with several video frames as the observation. Action prediction is based on the consistency of the scene context in the videos and most existing methods employ the appearance clues including the driving activity, human pose, structural relation of spatial context \cite{DBLP:journals/ijcv/WuWHLL21,DBLP:journals/ijcv/KongF22,DBLP:journals/tamd/LiLCL22,DBLP:journals/tcsv/ChenLSZ21,zheng2022parameter}. On the other hand, accident anticipation in driving scenarios need to exploit the strong context inconsistency in the normal and accident time window because of the severe camera motion and environment change during the accident events.

With the development of deep learning, accident anticipation formulations driven by large-scale video data have been established in recent years. One pioneering work proposed by Chan \emph{et al.} \cite{DBLP:conf/accv/ChanCXS16} model a Dynamic-Spatial-Attention Recurrent Neural Network (DSA-RNN) to correlate the temporal consistency of the tracklets of the road participants, and adopt an exponential loss for penalizing the delayed prediction. Following this work, an adaptive exponential loss function \cite{DBLP:conf/cvpr/SuzukiKAS18} is developed for earlier anticipation of accident and evaluated with a large-scale near-miss incident database. Similarly, the work \cite{DBLP:conf/cvpr/SuzukiKAS18} also follows the tracklet based input and infers the video context consistency by the Recurrent Neural Network (RNN) based modeling. Afterwards, many object-related accident anticipation methods are proposed, and commonly the RNN-based temporal correlation modeling and spatial interaction of objects (detected in advance) are the two primary insights in formulation. Based on this framework, Dynamic Spatial-Temporal Attention Network (DSTAN) \cite{DBLP:journals/tits/KarimLQY22} and Graph Convolutional Recurrent Neural Network (GCRNN) \cite{bao2020uncertainty} are proposed to infer the accident anticipation from the spatial-temporal relation learning perspective. Karim \emph{et al.} \cite{karim2022attention} propose an attention-guided multi-stream feature fusion for accident risk prediction. The discussed studies assume that there is an accurate detection for road participants, while this premise is difficult owing to the severe environment conditions, as well as the irregular shape of crashing objects.  

\subsection{Towards XAI for Accident Anticipation}
For safe-guaranteed driving, any AI model for accident prediction should consider its explainability \cite{DBLP:journals/ijcv/ZablockiBPC22}.  Karim \emph{et al.} \cite{DBLP:journals/corr/abs-2108-00273} propose a so-called explainable artificial intelligence for accident anticipation, which explores the activated heat map with Grad-CAM \cite{DBLP:conf/iccv/SelvarajuCDVPB17} method, and finds that the learned attention map is useful for boosting the explainability of the accident anticipation model. Bao and Kong \cite{DBLP:conf/iccv/Bao0K21} develop a Deep Reinforced accident anticipation with Visual Explanation (DRIVE) method with driver attention auxiliary on the DADA-2000 dataset \cite{DBLP:conf/itsc/FangYQXWL19,DBLP:journals/tits/FangYQXY22}, and claim that DRIVE is visually explainable. Actually, recent works for accident video analysis begin to focus the explainability, and a work in Singapore University of Technology and Design (SUTD) constructs a question answering benchmark with over 10k videos (consisting of surveillance and dashcam videos) for inferring the causal relation and QA problem in accident scenarios \cite{DBLP:conf/cvpr/XuHL21}. Closely related to this work, You and Han \cite{DBLP:conf/eccv/YouH20} specially investigate the causality recognition of accident scenarios, and build the semantic taxonomy of traffic accident. Besides, spatial-temporal traffic scene graph embedding is also another branch for explainable AI model for traffic scene understanding, and Malawade \emph{et al.} \cite{DBLP:journals/corr/abs-2111-06123} exploit the scene graph in early vehicle collision prediction. In addition, collision risk reasoning is important for accident anticipation, and Schoonbeek \emph{et al.} \cite{DBLP:conf/ivs/SchoonbeekPAD22} predict the collision risk with simulated video data and formulate a driving task-specific RiskNet by attentively masking the front vehicles for safe status prediction of the driving scene.

Based on these studies, we can observe that XAI has become a core direction for accident anticipation. In this work, we also follow such trend. Specifically, we will exploit the text description in the observation before the accident as well as the driver attention in driving situations as two important clues for modeling an explainable and cognitive accident prediction.

\subsection{Accident Prediction Datasets}
The community has realized the importance of accident prediction for safe driving, and some benchmarks have been released in recent years. In order to present a clear review, the characteristics of the available accident prediction datasets are presented in Table. \ref{tab1}. Chan \emph{et al.} \cite{DBLP:conf/accv/ChanCXS16} contribute to the DAD dataset, where each video clip is trimmed with 10 accident frames at the end of each clip. This setting is also adopted in the CCD Datasets \cite{bao2020uncertainty} with 50 frames for each clip. However, this setting is hard to be maintained in practical use and involves extra misleading prior that the accident occurs in the end of the clip. Therefore, some new benchmarks without trimming are released recently, such as A3D \cite{DBLP:conf/iros/YaoXWCA19} and DoTA \cite{yao2022dota}, which are used for unsupervised traffic accident detection \cite{DBLP:conf/iros/YaoXWCA19,yao2022dota,DBLP:journals/tits/FangQBYX22}. Similarly, we also construct the DADA-2000 dataset \cite{DBLP:journals/tits/FangYQXY22} with extra driver attention data, and inspired by our work, Eyecar \cite{DBLP:conf/iccv/BaeePK0OB21} is contributed for driver attention prediction problem in accident scenarios. Besides the dashcam videos for accident analysis, some researchers \cite{shah2018accident,DBLP:conf/cvpr/XuHL21} introduce traffic accident datasets in surveillance view, which will not be further investigated in this work considering the topic difference. 

 \begin{figure}[!t]
  \centering
 \includegraphics[width=\linewidth]{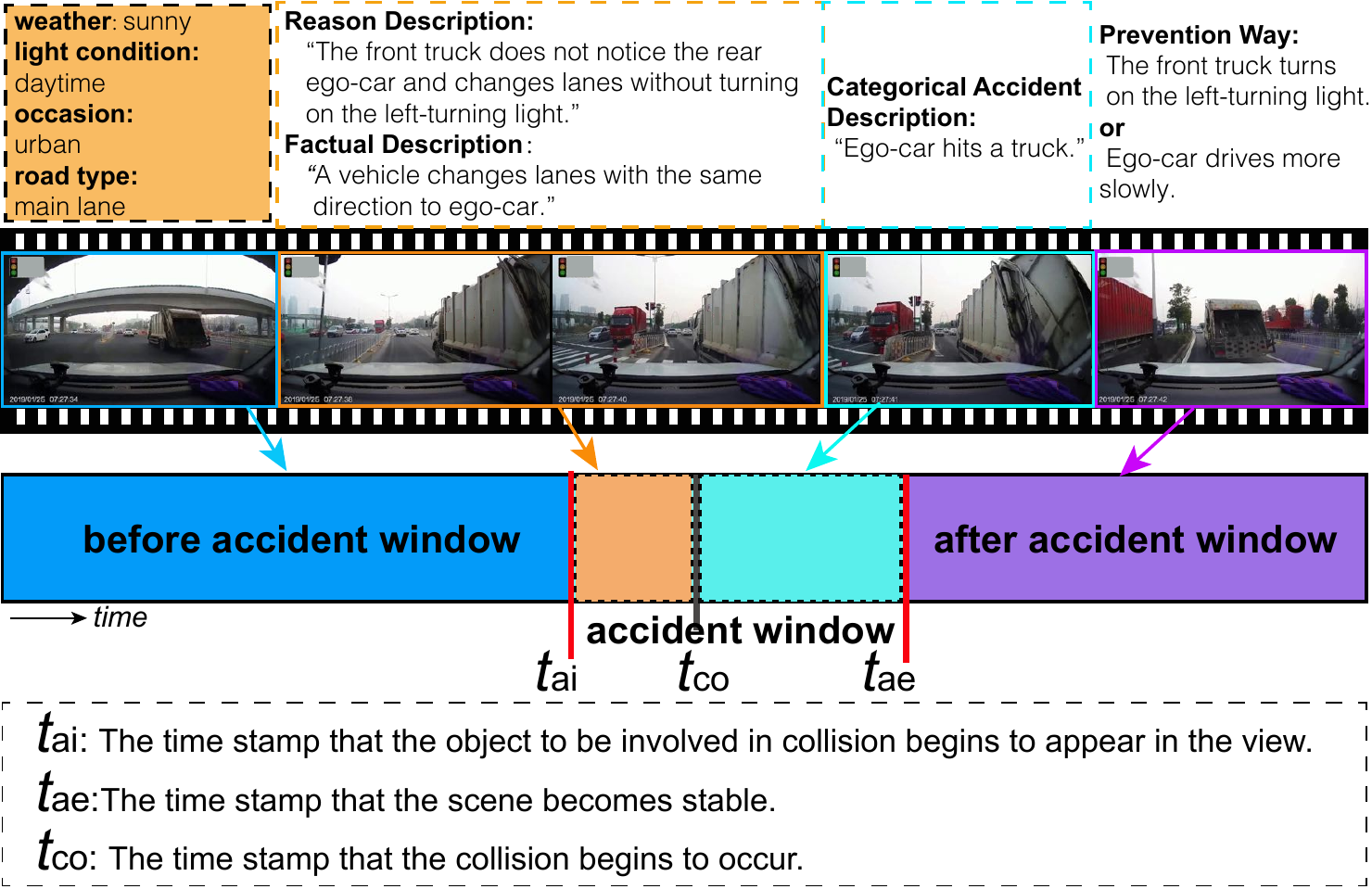}
  \caption{Illustration for different kinds of attribute annotations in CAP-DATA.}
  \label{fig2}
\end{figure}
    \begin{table*}[htpb]\scriptsize
  \centering
  \caption{Static-att statistic (videos) on weather condition, light condition (daytime:DT.; nighttime: NT.), occasion situations, and road types (main lane: m-lane; curve road: cur-rd; intersection: intsec; and T-road: T-rd). Note, the statistics are conducted on the accident videos.}
    \renewcommand{\arraystretch}{1.2}
     \setlength{\tabcolsep}{1.6mm}{
\begin{tabular}{c|cccc|cc|ccccc|ccccc}
\toprule[0.8pt]
 \multirow{2}[16]{*}{} & \multicolumn{4}{c|}{weather condition} & \multicolumn{2}{c|}{light condition}& \multicolumn{5}{c|}{occasion situations} & \multicolumn{5}{c}{road type}\\
\cmidrule{2-17}          & sunny &rainy&snowy& fogy &DT.& NT. &highway&urban &rural&mountain& tunnel &m-lane&cur-rd &intsec&T-rd& ramp\\
 \hline
CCD \cite{bao2020uncertainty} &1306& 61 & 14&0&1247& 134&148& 725 & 502&5&1&813& 94 & 258&210&6\\
A3D \cite{DBLP:conf/iros/YaoXWCA19} & 2990& 251 &474&42& 3340& 417& 225& 2458 &720&328&26& 2056& 190 &1100&363&48\\
DoTA \cite{yao2022dota} & 4920& 341 &313&12& 4943& 643& 617& 3656 &1148&145&20& 3033& 398 &1537&561&57\\
DADA-2000\cite{DBLP:journals/tits/FangYQXY22} & 1860& 130 &10&-& 1800& 200& 1420& 380 &180&-&20& 835& 99 &608&391&22\\
CAP-DATA & 10116& 761&793&57& 10463& 1264& 1082& 7563&2548&484&50& 6209& 685&3302&1410&121\\
  \hline
  \end{tabular}}
  \label{tab2}
  \end{table*}
  
\section{CAP-DATA Benchmark}
\label{ben}
The videos in CAP-DATA are collected from the publicly available accident datasets, and various video stream sites, such as Youtube, Bilibili, and Tencent, \emph{etc}. Among them, the CCD, A3D, DoTA, and our DADA-2000 datasets are used with further text annotation, careful anomaly window, and accident time stamp annotation. Besides, we also divide all the accident videos into 58 kinds of accident categories based on the occurrence relation of different road participants. Totally, \textbf{11,727} videos with \textbf{2.19 million} frames are collected and annotated. Based on the annotation attributes, CAP-DATA can support many useful tasks for accident inference, such as accident detection and prediction (Accident-Det/Pre), causal inference of accident (Accident-Causal), accident classification (Accident-Cla), text-video-based accident retrieval (Accident-Retri), and question answering in an accident (Accident-QA) of the driving scene. The time window with different annotations is illustrated in Fig. \ref{fig2}.

\subsection{Attributes and Statistics of CAP-DATA}
Attributes of CAP-DATA can be divided into three classes: frame-level time stamp for accident window (Frame-att), textual descriptions for accident inference (Textual-att), and the static driving scene attributes (Static-att). 

 \begin{figure}[htpb]
  \centering
 \includegraphics[width=\hsize]{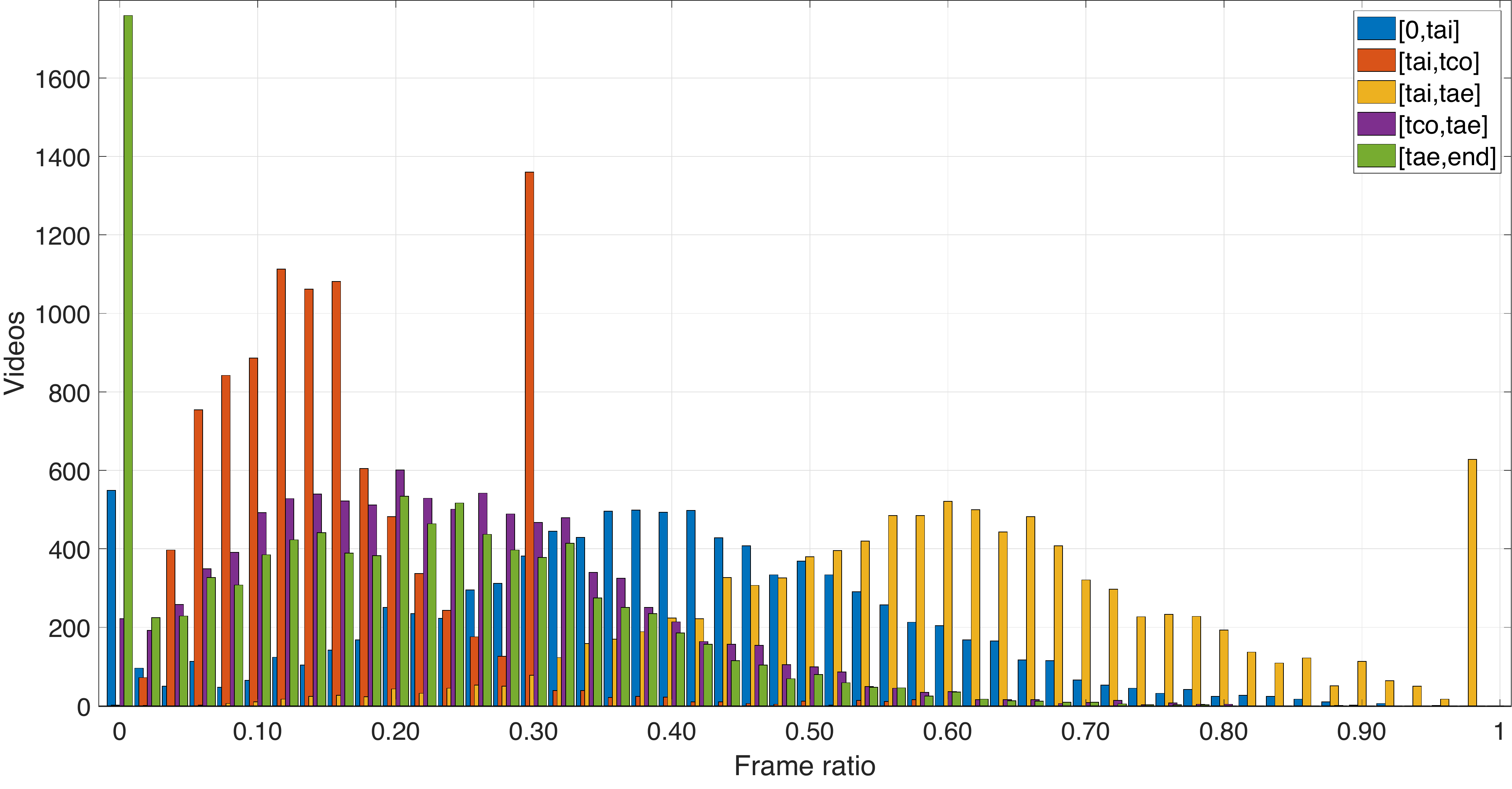}
  \caption{Frame ratios of the number of frames within the time window of [0,$t_{ai}$], [$t_{ai}$,$t_{co}$], [$t_{ai}$,$t_{ae}$], [$t_{co}$,$t_{ae}$], and [$t_{ae}$, end]. }
  \label{fig3}
\end{figure}
 \begin{figure}[htpb]
  \centering
 \includegraphics[width=0.8\hsize]{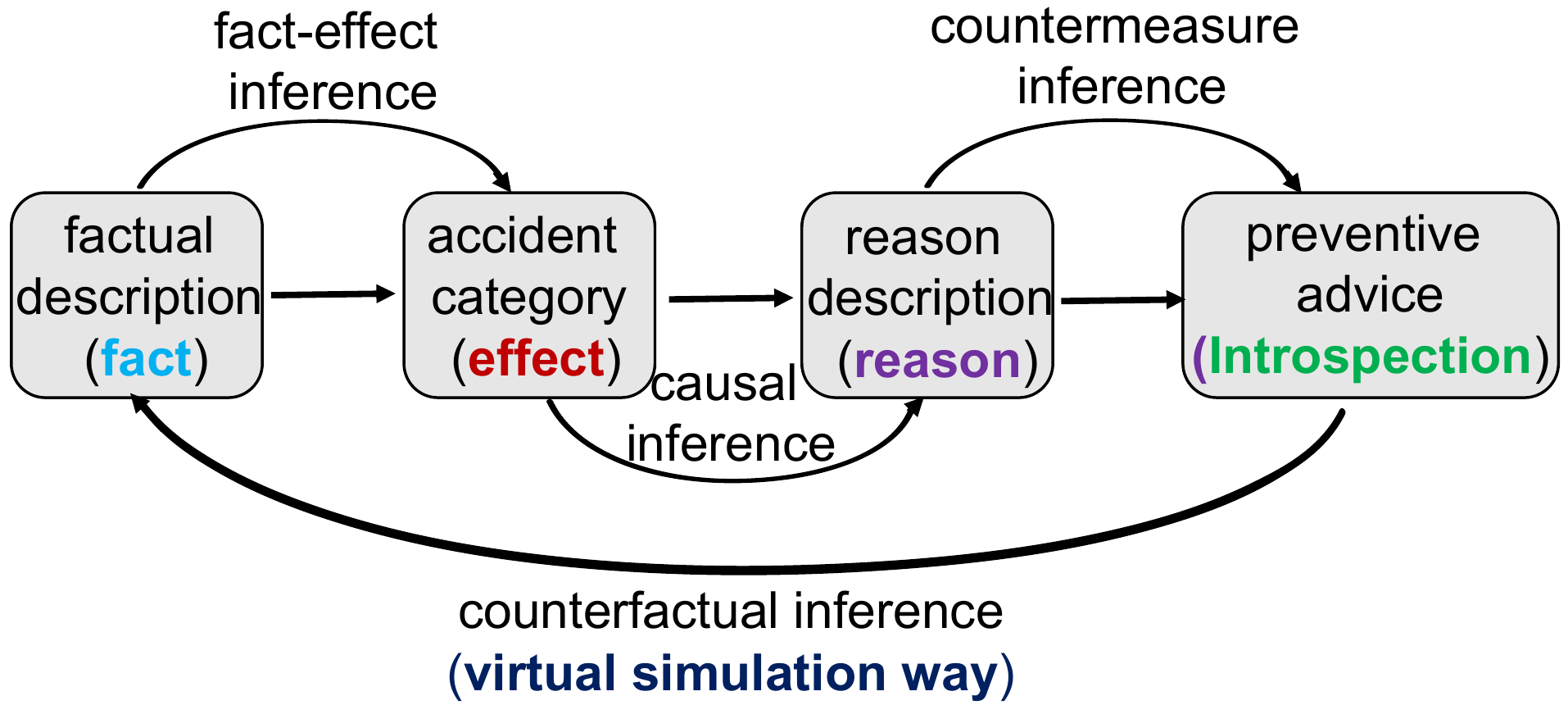}
  \caption{The linking cycle of fact-effect-reason-introspection descriptions.}
  \label{fig4}
\end{figure}

\textbf{Frame-att} contains the beginning time of accident window $t_{ai}$, the end time of accident window $t_{ae}$, and the time stamp $t_{co}$ where the collision begins to occur. The frame ratio distributions within different time windows of [0,$t_{ai}$], [$t_{ai}$,$t_{co}$], [$t_{ai}$,$t_{ae}$], [$t_{co}$,$t_{ae}$], and [$t_{ae}$, end] are shown in Fig. \ref{fig3}. It can be seen that, in many videos, the accident window (the crashing object appears in the field of view) starts from the first frame and many accidents end in the last frame. The accident window of [$t_{ai}$,$t_{ae}$] of most videos occupy half of the video length, which is useful for model training in accident prediction.
 \begin{figure}[!t]
  \centering
 \includegraphics[width=\hsize]{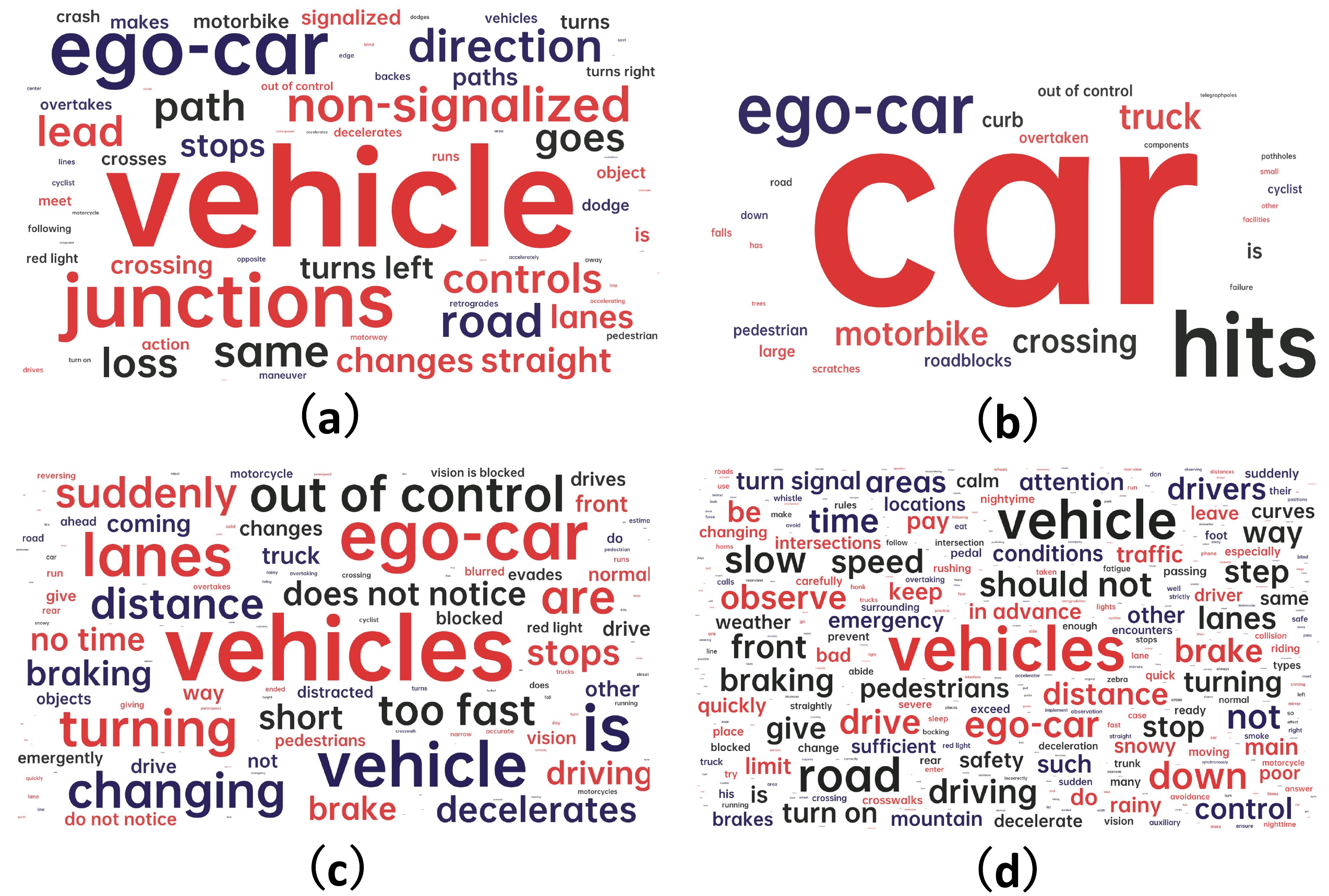}
  \caption{The keyword and phrase cloud for (a) fact, (b) effect, (c) reason, and (d) introspection descriptions on CAP-DATA with 11,727 videos, respectively.}
  \label{fig5}
\end{figure}

\begin{figure*}[!t]
  \centering
 \includegraphics[width=0.98\hsize]{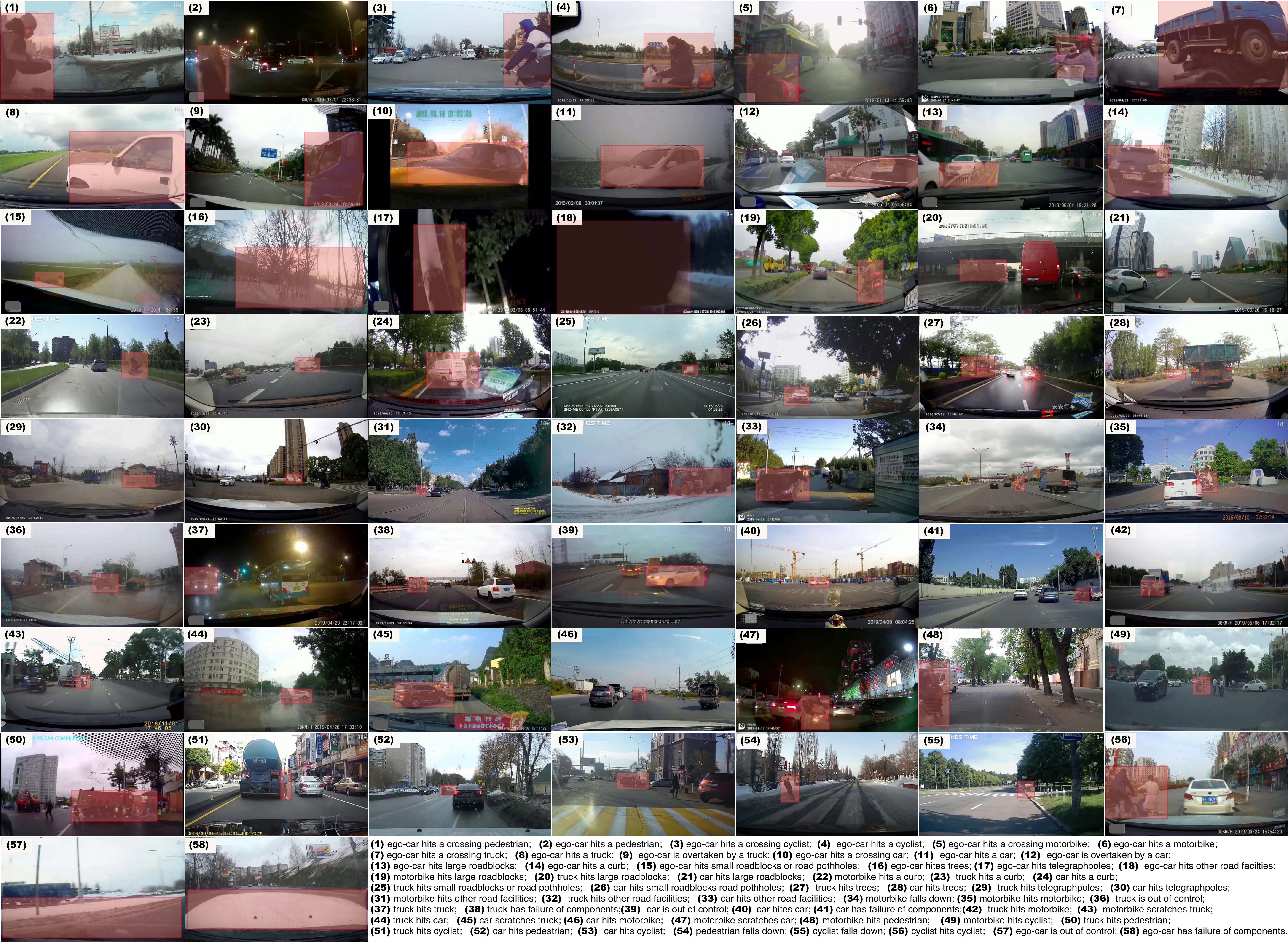}
  \caption{Some sample frames of 58 accident types, where the crash-object is marked by red bounding boxes. From these samples, we can observe that some accidents are challenging to be detected due to the small occurrence region.}
  \label{fig6}
\end{figure*}

\textbf{Textual-att} annotates the categorical accident description in the time window $[t_{co},t_{ae}]$, the reason description, and the factual description in the window of $[t_{ai},t_{co}]$, and the possible preventive advice for different kinds of accidents. Hence, there are four kinds of textual descriptions, \ie, accident category (\emph{effect}), accident reason description (\emph{reason}), the factual description before the accident (\emph{fact}), and the preventive advice (\emph{introspection}) for potential accidents. These descriptions form a cycle linking, as elaborated in Fig. \ref{fig4}, which paves the way for many promising accident inference tasks. Actually, the descriptions and the video sequence do not show a unique correlation, while each description sentence often correlates with many videos because of the co-occurrence. In this work, based on the road semantics, road user categories, and their movement actions, we annotated $72$ sentences for factual description, $58$ sentences for accident category description, and $110$ sentences for accident reason and introspection description, where the frame samples of $58$ accident categories are exhibited in Fig. \ref{fig6}. Specially, the reason description is annotated from the perspective of different types of road participants with primary responsibility for accidents, such as the examples of \emph{``a truck drives in speeding"}, \emph{``a vehicle overtakes when turning"}, and \emph{``a pedestrian stays on the motorway", etc}.

  \begin{figure*}[htpb]
  \centering
 \includegraphics[width=\hsize]{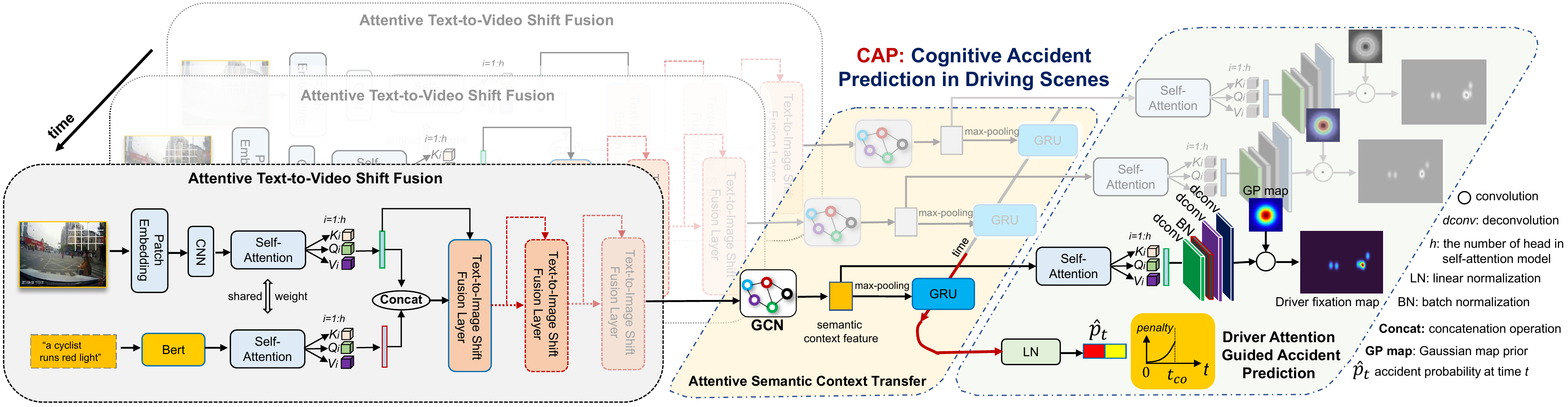}
  \caption{The pipeline of CAP. Given a video frame and the video-level factual text description before the accident, \textcolor{blue}{Attentive Text-to-Video Shift Fusion} module aims to learn the coherent semantics in video frame and text, where several Text-to-Image Shift Fusion Layers (T2I-SFLayer) are involved. T2I-SFLayer can be added repeatedly and the embedding after it is fed into the spatial \emph{GCN} model to extract the semantic context feature. With the temporal \emph{GRU}, the \textcolor{blue}{Attentive Semantic Context Transfer} module is modeled to serve the accident prediction score computation. In each time, the driver fixation map is also reconstructed by the self-attention of the semantic context feature, which aims to fulfill a \textcolor{blue}{Driver Attention Guided Accident Prediction}.}
  \label{fig7}
\end{figure*}
Fig. \ref{fig5} exhibits the word or phrase cloud for \emph{fact}, \emph{effect}, \emph{reason}, and \emph{introspection} description. From the text cloud, we can see that the accident category description (effect) contains minimal vocabulary and the preventive advice (introspection) has the most complex set of phrases. We can observe the difference between these types of descriptions from the core phrase or words, such as the ``\emph{junctions}", ``\emph{hits}", ``\emph{does not notice}", and ``\emph{should not}", respectively. It is worth noting that for the counterfactual inference between the strategy and fact description, it is impossible to change the past in real world, and virtual simulation may be a good alternative and will be investigated in the future.

\textbf{Static-att} labels the weather conditions, light conditions, occasions, and road types of driving scenes, where these attributes of CAP-DATA and other related datasets are presented in Table. \ref{tab2}. Most accidents occur in urban scenes with sunny daytime, while other accidents in severe conditions are hard samples. In addition, there are many samples of main lanes and intersections. In the experiments, we will evaluate the performance of different static attributes.

\subsection{Training and Testing Split}
CAP-DATA is rather large for performance evaluation. In order to validate the model feasibility efficiently and effectively for future research in this field, we provide a MINI benchmark (\textbf{MINI-Train-Test}) and a FULL benchmark (\textbf{FULL-Train-Test}) for training and testing. Because only DADA-2000 has the driver attention data for the training of the CAP model in this work, the MINI-Train-Test selects about half of DADA-2000, \ie, 1000 sequences for training and testing, where the setting is the same as the other works \cite{DBLP:conf/iccv/Bao0K21,DBLP:journals/tits/FangYQXY22}. In addition, for effective and adequate evaluation, FULL-Train-Test adopts the whole CAP-DATA for performance evaluation, where the whole sequences meeting the training condition in DADA-2000 are used for training, and the sequences meeting the testing condition in 9,727 sequences are adopted for testing. FULL-Train-Test needs huge and exhaustive work but can provide a full-portrait evaluation of the methods in this field. The training and testing setting will be explained in the experimental details.

\section{CAP Approach}
\label{method}
Given the observation of video frames, accident prediction aims to maximize the Time-to-Accident (TTA) $\tau$, where $\tau=max\{0,t_{ai}-t_{a}\}$, and $t_{a}$ denotes the first time stamp where the predicted accident probability $p_t$ is larger than a pre-defined threshold (\eg, 0.5 commonly). See Fig.\ref{fig2} for a clear illustration of the accident window, $t_{ai}$ is the beginning time of the accident.
In other words, accident prediction prefers an early awareness of $t_{a}$ before the accident. In this work, during the training of the model, we not only input the video frames in the observation but also the factual text description before the accident is taken. This setting aims to learn the coherent semantics in video and text and help to find useful semantics in accident prediction. Certainly, factual text description is not available in practical use, we thus adopt it in the training phase and design an \emph{attentive text-to-video shift fusion module} which aims to shift the text information to the vision path for maintaining the multimodality inference ability when testing the trained model with \textbf{only} the video data. 

Fig. \ref{fig7} shows the pipeline of the proposed CAP model, which consists of the \emph{attentive text-to-video shift fusion module}, an \emph{attentive semantic context transfer module}, and a \emph{driver attention guided accident prediction module}.
The video frame and the text description are encoded with two self-attention pathways initially and then fused together with the Text-to-Image Shift Fusion Layer (T2I-SFLayer). 
Given a video observation $\mathcal{V}_{1:t}=\{I_1,...,I_{t}\}$ and its factual text description $\mathcal{T}_{1:t}$, the output of the model CAP is defined as:
\begin{equation}
\hat{p_t} = \text{CAP}(\mathcal{V}_{1:t},\mathcal{T}_{1:t}),
\label{eq:1}
\end{equation}
where $\hat{p_t}$ is the accident probability at time $t$.

Similarly, in order to guide the semantic learning for critical regions in an accident situation, this work introduces the driver fixation map in the inference model of CAP. We predict the driver fixation map $\hat{{\bf{D}}}_t$ by the intermediate feature with self-attention.
By minimizing the predicted driver fixation map $\hat{{\bf{D}}}_t$ with the ground-truth ${\bf{D}}_t$, we aim to involve the human-inspired visual attention mechanism for accident prediction.

\subsection{Attentive Text-to-Video Shift Fusion}
\label{at2vsf}
In this work, we introduce the text description in the video during the training phase to assist the semantic learning of video-based accident prediction in driving scenes. Specially, we design an attentive text-to-video shift fusion model to learn the coherent semantics in accident videos in the training phase and shift the text information into video representations, so as to adapt to the testing stage without text description. 

\subsubsection{Self-Attentive Vision and Text Embedding}
Given the video frame $I_t\in \mathbb{R}^{224\times 224\times 3}$ and the text description $\mathcal{T}_t$,  they are encoded by the Patch Embedding and pre-trained BERT \cite{zhuang2021robustly} model, respectively. Patch Embedding is fulfilled by partitioning the video frame into $N\times N$ patches, and each patch is encoded by a 2D convolution with the kernel size of $16\times16$ with the stride of 16. Therefore, the patch width is $N=\frac{224}{16}=14$, and encoded video frame embedding is denoted as $M_t\in \mathbb{R}^{14\times 14\times 768}$ ($16\times16\times3=768$). In order to reduce the computational burden, $M_t^I$ is downsampled to ${7\times 7\times m}$ with a $1\times 1$ convolution, where $m$ is a hyper-parameter that denotes the dimension after downsampling convolution and is set as $m=120$ for all experiments. Besides, text description $\mathcal{T}_{1:t}$ is encoded by a pre-trained BERT model, and the output of the text embedding in this work is denoted as $M^T_t\in \mathbb{R}^{n_{T}\times m}$, where $n_{T}$ is also a hyper-parameter which is determined by the max length of the input text and is set as 15 in this work.

We next feed the embeddings of the video frame and text description into a Multi-Head Self-Attention (MHSA) model \cite{DBLP:conf/nips/VaswaniSPUJGKP17} with shared weights to model the correlation of the visual patches or text words. In order to make the video frame embedding feasible for MHSA, we flatten $M^I_t$ into a $49\times 120$ matrix. The number of heads in MHSA is set as 8. Consequently, the number of tokens in vision and text pathways are $49$ and $15$, respectively. The outputs of MHSA in vision and text embeddings are denoted as ${\bf{O}}^I_t$ and ${\bf{O}}^T_t$ at time $t$, respectively, which have the same dimension with $M^I_t$ or $M^T_t$. ${\bf{O}}^I_t$ and ${\bf{O}}^T_t$ are then fed into the following T2I-SFLayer.

\subsubsection{T2I-SFLayer}
\begin{figure}[!t]
  \centering
 \includegraphics[width=\hsize]{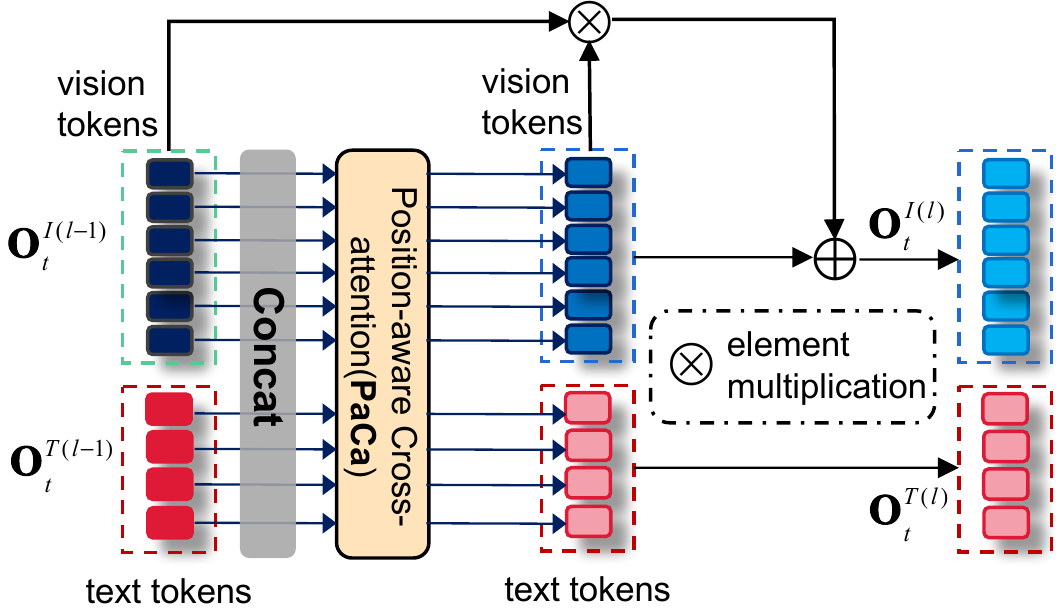}
  \caption{Overview of T2I-SFLayer.  }
  \label{fig8}
\end{figure}

T2I-SFLayer fulfills the shift fusion of text tokens ${\bf{O}}^T_t$ to vision tokens ${\bf{O}}^I_t$ after a self-attention model. In order to provide an attentive fusion, we adopt a Position-aware Cross-attention (PaCa) to combine the text and vision modalities, which aims to maintain the token position after the fusion layer and explore the relation of local and global token representations. In particular, considering that there is no available text information in practical use, we design a cross-layer shift operation of vision tokens to enhance the learning of vision information, as shown in Fig. \ref{fig8}. Assume ${\bf{O}}^T_t$ and ${\bf{O}}^I_t$ are the $0^{th}$ layer of tokens in the fusion process. We denote ${\bf{O}}^{T(l)}_t$ and ${\bf{O}}^{I(l)}_t$ as the output text and vision tokens after the $l^{th}$ layer of PaCa. T2I-SFLayer is fulfilled by:
\begin{equation}
\begin{aligned}
{\bf{O}}^{F(l-1)}_t=\text{Concat}({\bf{O}}^{T(l-1)}_t,{\bf{O}}^{I(l-1)}_t),\\
[{\bf{O}}^{T(l)}_t,{\bf{O}}^{I(l)}_t]=\text{PaCa}({\bf{O}}^{F(l-1)}_t),\\
{\bf{O}}^{I(l)}_t={\bf{O}}^{I(l)}_t\otimes({\mathbf{1}}+{\bf{O}}^{I(l-1)}_t),
\end{aligned}
\end{equation}
where $\otimes$ means element multiplication, $\text{Concat}(\cdot)$ denotes the concatenation operation, and ${\mathbf{1}}$ is a vector with the same shape as the vision tokens. 

\textbf{PaCa}: In order to involve the local and global relation in PaCa, we partition the fused token matrix ${\bf{O}}^{F(l-1)}_t\in \mathbb{R}^{64\times m}$ ($m=120$) into $h$ local heads ${\bf{o}}^{F(l-1)}_{t,j}\in \mathbb{R}^{64\times \frac{m}{h}}$($j\in\{1,...,h\}$), and the global token representation is inputted to a one-layer of MLP ($LN (120) \rightarrow relu \rightarrow droupout\rightarrow LN (240) \rightarrow droupout$) to obtain the global summarization ${\bf{\hat{B}}}^{F(l-1)}_t$ with the same dimension ($\frac{m}{h}$) to local head. $PE$ is the position embedding. $h$ is typically set as 8. PaCa($\cdot$) is fulfilled by: 
\begin{equation}\small
\begin{aligned}
&{\bf{\hat{o}}}^{F(l-1)}_j=\text{MHSA}(Q_f({\bf{o}}^{F(l-1)}_{t,j}+{PE}),K_f({\bf{\hat{B}}}^{F(l-1)}_t+{PE}),\\ &\verb'        'V_f({\bf{\hat{B}}}^{F(l-1)}_t)),\\
&{\bf{O}}^{I(l)}=\text{Concat}({\bf{\hat{o}}}^{F(l-1)}_{t,1}[1:49,:],...,{\bf{\hat{o}}}^{F(l-1)}_{t,h}[1:49,:])\\
&{\bf{O}}^{T(l)}=\text{Concat}({\bf{\hat{o}}}^{F(l-1)}_{t,1}[50:64,:],...,{\bf{\hat{o}}}^{F(l-1)}_{t,h}[50:64,:])
 \end{aligned}
\end{equation}
where $K_f$, $Q_f$, and $V_f$ are the key, query and value representation with the $LN$ (linear layer) on $({\bf{o}}^{F(l-1)}_{t,j}+{PE})$, $({\bf{\hat{B}}}^{F(l-1)}_t+{PE})$, and ${\bf{\hat{B}}}^{F(l-1)}_t$, respectively. 

Actually, after each PaCa layer, the text and vision tokens have attentively correlated and each token has the related text and vision information. With the Position Embedding ($PE$), the cross-layer shift operation of vision tokens enhances the vision semantic learning and assigns coherent text information in the learning process, which fulfills an attentive text-to-video shift fusion process. 

\subsection{Attentive Semantic Context Transfer}
\label{asct}
After the Attentive Text-to-Video Shift Fusion, the fused cross-modal feature is denoted as ${\bf{O}}_t^F \in \mathbb{R}^{(15+49)\times m}$, which has $64$ tokens, and each token represents the specific activated semantic information. Then, this work explores the semantic context in the driving scene by one-layer of graph convolution over the fused cross-modal feature ${\bf{O}}_t^F$, and defined as:
\begin{equation}
{\bf{S}}_t=\text{GCN}({\bf{O}}_t^F,{\bf{A}}_t),
\end{equation}
where ${\bf{S}}_t$ denotes the $64\times K$ semantic context feature matrix and each row specifies a node representation, where $K$ is set as 512. ${\bf{A}}_t$ is the adjacent matrix of ${\bf{O}}_t^F$ and is computed by Euclidean distance across the node representations.

It is worth noting that the semantic context feature matrix ${\bf{S}}_t$ is fed into the following two pathways. One is used to infer the video-level accident prediction and another is adopted to reconstruct the frame-level driver attention map. 

Because the accident prediction infers the temporal correlation of the semantic context features, we leverage the temporal memory learning with Gated Recurrent Unit (GRU) to correlate the semantic context feature over time. The input of GRU is obtained by a max-pooling operation on ${\bf{S}}_t$ in the node dimension and is denoted as ${\bf{s}}_t\in \mathbb{R}^{1\times 512}$. Then, the GRU module is fulfilled by:
\begin{equation}
{\bf{H}}_{t}=\text{GRU}([{\bf{s}}_{1},...,{\bf{s}}_{t}], {\bf{H}}_{t-1}, \Omega_s),
\end{equation}
where 
${\bf{H}}_{t-1}$ and $\Omega_s$ are the hidden state representation at $t^{th}$ recurrent step and the learnable weight parameter of GRU, respectively.  ${\bf{H}}_{0}$ is initialized with the dimension of $k\times 1 \times 256$. From the temporal GRU, the accident score $\hat{p}_t$ is computed by decoding the hidden state representation ${\bf{H}}_{t}$ with two fully connected layers ($fc$) with the parameters of $\theta_1$ and $\theta_2$ and followed a \text{softmax} function, which is defined as:
\begin{equation}
\hat{p}_t=\text{softmax}(fc(fc({\bf{H}}_{t};\theta_1);\theta_2)).
\end{equation}
\subsection{Driver Attention Guided Accident Prediction}
\label{dagap}

\subsubsection{Driver Attention Map Reconstruction}
The semantic context feature matrix ${\bf{S}}_t$ at time $t$ is re-used for the driver attention map reconstruction, which aims to provide a traction role for the core semantic learning in accident prediction. This strategy is inspired by our previous finding \cite{DBLP:conf/itsc/FangYQXWL19}. In order to find the core node representation in the semantic context feature matrix, we again adopt a similar MHSA with the vision embedding in this work.

Assume the output after the self-attention model is ${\bf{\hat{S}}}_t\in \mathbb{R}^{64\times 512}$, it is decoded as ${{\bf{D}}^{tmp}}_t\in \mathbb{R}^{64\times 64}$ by an interleaved layers of deconvolution and batch normalization, and the structure is $\{dconv(3\times3,64) \rightarrow BN \rightarrow relu \rightarrow upsampling(\times 4) \rightarrow dconv(3\times3,16) \rightarrow relu\rightarrow dconv(5\times5,1)\rightarrow relu\}$, where $dconv$ (kernel size, channels) serves the deconvolution. The decoded representation ${\bf{D}}^{tmp}_t$ is filtered as ${\bf{\hat{D}}}_t$ by a convolution operation with a Gaussian kernel $\text{Gauss}(w, \sigma)$, where $w$ is the Gaussian kernel with the size of $3\times 3$, and $\sigma$ denotes the variation with the value of $1.5$.

For driver attention map reconstruction, the loss function $\mathcal{L}_d$ is defined as the summarization of Kullback-Leibler divergence (KLdiv) distance over all frames in the training samples:
\begin{equation}
\begin{aligned}
\mathcal{L}_d=\sum_{sn}^{SN}[\sum_{t=1}^{N_I}\sum_i {{\bf{D}}_t(i)} \text{log}(\epsilon+\frac{{\bf{D}}_t(i)}{\epsilon+{\bf{\hat{D}}}_t(i)})],
\end{aligned}
\end{equation}
where $SN$ means the number of clip samples in each batch, $i$ is the pixel index in ground-truth ${\bf{D}}_t$, and $\epsilon$ is a small constant (0.0001) for ensuring numerical stability. $N_I$ is the number of frames in each clip sample.
\subsubsection{Accident Prediction}
Accident prediction in this work is denoted as a classification problem and we sample the positive (with accident) and negative (without accident) video clips for model training. For each long accident video, the positive samples are obtained by taking the end frame of the sample in the accident window (\ie, the frames in $[t_{ai},t_{co}]$), and the negative samples are generated from the time window $[0,t_{ai}]$, which is the same as the work of DRIVE \cite{DBLP:conf/iccv/Bao0K21}. For accident prediction, we apply different penalties to the negative and positive samples. The positive samples imply an earliness of prediction when meeting the actual accident. The loss function for accident prediction is defined as:
\begin{equation}
\begin{aligned}
\mathcal{L}_a=-\sum_{sn}^{SN}[\sum_{t=1}^{N_I} (1-y_{t})\text{log}(1-\hat{p}_t)\\
+\sum_{t=1}^{N_I} y_{t}e^{-\text{max}(0,\frac{t_{ai}-t}{r})}\text{log}(\hat{p}_t)],
\end{aligned}
\end{equation}
where $r$ is the frame rate. If the index $sn$ points to a negative sample, then $y_{t}=0$, and the cross entropy loss is used. Otherwise, $y_{t}=1$ and an exponential loss function \cite{DBLP:conf/cvpr/SuzukiKAS18} is adopted. $t_{ai} - t$ is the time interval $\tau$ between the latest observation with the beginning time of the actual collision. $\hat{p}_t$ are the predicted accident probability at frame $t$.

\subsubsection{Multi-task Optimization}
In this work, the driver attention map reconstruction and traffic accident prediction form a multi-task learning problem, and the loss function is:
\begin{equation}
\mathcal{L}=\mathcal{L}_d+\lambda\mathcal{L}_a, 
\end{equation}
where $\lambda$ is the parameter for balancing the error gap, and is set as 5 in all experiments. For the optimization, we back-propagate the gradient in sample-level, \ie, the weights of the model are updated when one training video sample is traversed completely. Because these two tasks have large gap, where driver attention reconstruction needs at frame-level optimization, accident prediction is a video-level optimization task. In order to fulfill a joint optimization, we set different learning rates on different modules, to be described in the experimental details. 

\section{Experiments}
\label{experiments}
\subsection{Datasets and Evaluation Metrics}
In this work, we first construct the experiments on our CAP-DATA dataset with the MINI-Train-Test and FULL-Train-Test evaluations. Then, we also evaluate the performance of the CCD dataset \cite{bao2020uncertainty} to present a fair comparison with other state-of-the-art methods. CCD dataset \cite{bao2020uncertainty} contains 1,500 dashcam accident videos. Each video is trimmed into 50 frames with a totally of 5 seconds long. For each accident, it is placed in the last 2 seconds of each accident video. Similar to the setting of CCD, 900 accident videos are chosen for the performance evaluation in our work.

\emph{Evaluation Metrics:} To verify the model effectiveness, we deploy the same metrics used in previous works \cite{bao2020uncertainty,DBLP:conf/iccv/Bao0K21}, \ie, Average Precision (AP) \cite{bao2020uncertainty}, and Time-to-Accident (TTA) \cite{DBLP:conf/iccv/Bao0K21} which includes the variants of TTA$_{0.5}$ and mTTA. 
TTA$_{0.5}$ is computed by setting the accident score threshold in each video as 0.5 to evaluate the earliness of a positive prediction, and mTTA denotes the mean TTA which is computed by the mean value of TTA with the changing of accident score threshold from 0 to 1. AP evaluates the average accuracy with different thresholds of precision and recall rate. In this work, we also evaluate the methods based on video-level Area under ROC curve (AUC) \cite{DBLP:conf/iccv/Bao0K21}. AUC is a metric for evaluating the detection or classification, which is adopted for checking the prediction performance with the output of the occurrence probability of future accidents. For these metrics,  a larger value means better performance.   

Besides, we also check the performance of driver attention reconstruction with the same metrics of KLdiv, Correlation coefficient (CC), Similarity Metric (SIM), and center-biased area under ROC curve (s-AUC) in \cite{DBLP:journals/tits/FangYQXY22} for evaluating the interactive role with accident prediction. Except KLdiv, CC, SIM, and s-AUC pursue a large value.

 \begin{figure}[!t]
  \centering
 \includegraphics[width=0.89\linewidth]{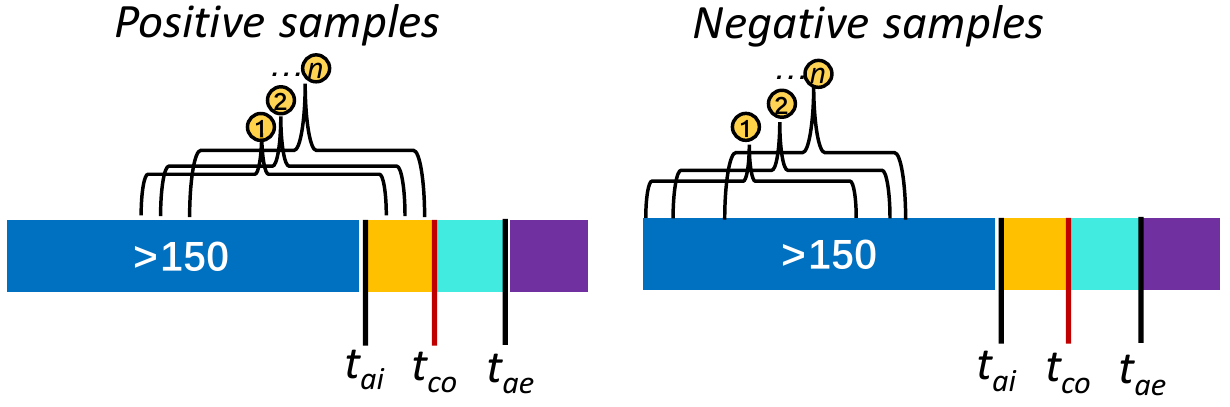}
  \caption{The sampling strategy for the positive and negative samples on the training set of MINI-Train and FULL-Train, where $t_{co}$ is the time that the collision begins to occur, and [$t_{ai}$, $t_{ae}$] is the accident window.}
  \label{fig9}
\end{figure}
\begin{table}[!t]\small
  \centering
  \caption{The number of training and testing samples in MINI-Train-Test and FULL-Train-Test evaluation.}
    \renewcommand{\arraystretch}{1.2}
\begin{tabular}{c|cccc}
\toprule[0.8pt]
   \multirow{2}[4]{*}{Eval.} & \multicolumn{4}{c}{MINI-Train-Test}\\
\cmidrule{2-5}          & Raw Videos &  &Positive & Negative \\
\hline
Train  &512&$\rightarrow$ sampling$\rightarrow$& 1778&1534\\
Test&168&$\rightarrow$ sampling$\rightarrow$&608&207\\
\hline\hline
 \multirow{2}[4]{*}{}&\multicolumn{4}{c}{FULL-Train-Test}\\
    \cline{2-5}  & Raw Videos &  &Positive & Negative\\
    \hline
  Train &1020&$\rightarrow$ sampling$\rightarrow$&1530&1334\\
Test-5s &1281&$\rightarrow$ sampling$\rightarrow$ &1281&136\\
Test-4s &2044&$\rightarrow$ sampling$\rightarrow$ &2044&216\\
Test-2s &4775&$\rightarrow$ sampling$\rightarrow$ &4775&2469\\
 \hline
  \end{tabular}
  \label{tab3}
  \end{table}

  \begin{table*}[!t]\small
  \centering
  \caption{Testing performance evaluation on the role of textual description and T2I-SFLayers on the MINI-Test set, where the best value of each metric in different number of T2I-SFLayers is marked in \textbf{bold} font.}
    \renewcommand{\arraystretch}{1.2}
         \setlength{\tabcolsep}{1.6mm}{
\begin{tabular}{ccccc|cccc|cccc}
\toprule[0.8pt]
   \multirow{2}[4]{*}{Baselines} & \multicolumn{4}{c|}{One T2I-SFLayer} & \multicolumn{4}{c|}{Two T2I-SFLayers} & \multicolumn{4}{c}{Three T2I-SFLayers}\\
\cmidrule{2-13}          & AP$\uparrow$ &AUC$\uparrow$  & TTA$_{0.5}\uparrow$ & mTTA$\uparrow$ & AP$\uparrow$ &AUC$\uparrow$  & TTA$_{0.5}\uparrow$&mTTA$\uparrow$ & AP$\uparrow$ &AUC$\uparrow$  & TTA$_{0.5}\uparrow$&mTTA$\uparrow$ \\
\hline
CAP$_{\text{-Text+Att}}$  &0.772&0.766& 3.847&4.241&\textbf{0.758}&0.822& 3.877&4.347&0.744&\textbf{0.833}& \textbf{4.020} &\textbf{4.497}\\
CAP$_{\text{+Text+Att}}$ &0.755&0.684& \textbf{4.076}&\textbf{4.648}&0.739&0.820& 4.063&4.483&0.753&0.819& 3.951 & 4.464\\
 \hline
  \end{tabular}}
  \label{tab4}
  \end{table*}

 \subsection{Implementation Details}
The performance evaluation is conducted on the MINI-Test and FULL-Test set of CAP-DATA, and the hardware configuration contains an Nvidia RTX 2080Ti GPU and 64G RAM. The running efficiency is about 15fps. For the MINI-Train-Test, with the same setting of DRIVE \cite{DBLP:conf/iccv/Bao0K21}, we take 512 and 168 raw video sequences for training and testing. As for the FULL-Train-Test evaluation, we use 1,020 raw videos (each video has over 150 frames) in DADA-2000 for training and other video sequences without the driver attention data for testing. Fig. \ref{fig9} shows the sampling strategy of positive samples and negative samples in the training dataset of MINI-Train and FULL-Train. For MINI-Train, because of the limited training samples, we randomly sample the temporal window with the end time of $t_{co}$ (with the length of 150 frames) in raw accident videos, and the positive and negative samples are defined by checking whether the collision frame at time $t_{ai}$ is contained in the sampled window or not. Consequently, the positive and negative samples in MINI-Train-Test have many overlapping frames. The sampling strategy is the same as in DRIVE \cite{DBLP:conf/iccv/Bao0K21}. As for the Full-Train-Test, we have more raw training videos, and we sample the positive and negative samples with fewer overlapping frames. Total, the number of the training and testing samples in MINI-Train-Test and Full-Train-Test are shown in Table. \ref{tab3}. Notably, for the testing set in Full-Train-Test evaluation, we have three cases of 5s-prediction, 4s-prediction, and 2s-prediction. In different cases, because of the imbalanced frame length for all video sequences, the number of negative and positive samples that we can obtain by the sampling strategy is not the same.

In order to optimize the accident prediction and driver attention map reconstruction simultaneously, we set different learning rates for the Self-Attention block, T2I-SFLayers, GRU, and the decoding module of driver attention map as $10^{-6}$, $10^{-6}$, $10^{-5}$, and $10^{-4}$, respectively. All experiments are trained with 10 epochs, and the batch size is set as 2.

\subsection{Ablation Studies}
As for the text role, based on the latest survey for multimodal learning with transformers \cite{DBLP:journals/corr/abs-2206-06488}, multimodal input is better than one modal version in many applications. Therefore, we mainly open two parts of ablation studies to discuss the role of T2I-SFLayers and the role of driver attention. The MINI-Test is adopted here for the ablation studies.
\subsubsection{Does the module of T2I-SFLayer work for transferring the text information to the vision path during the testing phase?}
As aforementioned, factual text descriptions before the accident can provide guidance for finding the object involved in the crash efficiently under different environments. Therefore, during testing, we evaluate whether the trained model has the ability for transferring the text information to the vision path with the increase of T2I-SFLayers effectively. This question is verified by checking the performance difference for the testing phase with or without the text description. In order to maintain the multimodality input mode, the model version without text is fulfilled by changing the original text description in training to a consistent description of ``\emph{a frame of $\{ \}$}" in testing. Consequently, we compare the model with text input (\textbf{CAP$_{\text{+Text+Att}}$}) \footnote{Notably, we have annotated the text information for training and testing data. Here, the \textbf{CAP$_{\text{+Text+Att}}$} is fulfilled by inputting the annotated factual text information in comparison during testing. It is the same in the following analysis for the model noted the ``+Text" subscript.} and the model without text input (\textbf{CAP$_{\text{-Text+Att}}$}) to check text influence. In addition, the text-to-image shift fusion is fulfilled by T2I-SFLayer. From Table. \ref{tab4}, we observe that vision information benefits the AP and AUC values. In addition, with the increase of T2I-SFLayers, the difference for the TTA metrics becomes smaller, and more T2I-SFLayers increase the values of [TTA$_{0.5}$, mTTA] of \textbf{CAP$_{\text{-Text+Att}}$}  from [$3.847$, $4.241$] to [$4.020$, $4.497$]. Therefore, T2I-SFLayer works for the testing only with the video data.

\begin{table}[!t]\small
  \centering
  \caption{The performance comparison for the driver attention role on the MINI-Test under the setting of three T2I-SFLayers without the textual description.}
    \renewcommand{\arraystretch}{1.2}
\begin{tabular}{ccccc}
\toprule[0.8pt]
Baselines  & AP$\uparrow$ &AUC$\uparrow$  &TTA $_{0.5}$$\uparrow$&mTTA$\uparrow$\\
\hline
DRIVE \cite{DBLP:conf/iccv/Bao0K21}&0.690&0.727& 3.657& 4.259\\
\hline
CAP$_{\text{-Text-Att+3}}$  &0.701&0.774& 3.742& 4.284\\
CAP$_{\text{-Text+Att+3}}$  &\textbf{0.744}&\textbf{0.833}& \textbf{4.020}& \textbf{4.497}\\
 \hline
  \end{tabular}
  \label{tab5}
  \end{table}
 
\subsubsection{What about the interactive role between accident prediction and driver attention reconstruction?}
Beside the text description, driver attention is another clue for finding the core semantics for accident prediction. Here, we evaluate the cross-task promotion role in training between driver attention map reconstruction and accident prediction. Based on the evaluation of T2I-SFLayers, we select the version of three T2I-SFLayers without text input here for comparison. The trained baselines for accident prediction are termed as  CAP$_{\text{-Text+Att+3}}$ and CAP$_{\text{-Text-Att+3}}$ for with and without driver attention reconstruction. The results are presented in Table. \ref{tab5}, and indicate that the driver attention plays the manifested role in accident prediction. We also include the results of DRIVE \cite{DBLP:conf/iccv/Bao0K21} here, and it shows that our method could improve performance significantly. 
 \begin{table}[htpb]\small
  \centering
  \caption{Driver attention map results on the MINI-Test set.}
    \renewcommand{\arraystretch}{1.2}
         \setlength{\tabcolsep}{1.6mm}{
\begin{tabular}{ccccc}
\toprule[0.8pt]
Baselines  & KLdiv$\downarrow$ &CC$\uparrow$  &SIM $\uparrow$  &s-AUC$\uparrow$ \\
\hline
BDDA \cite{DBLP:conf/accv/XiaZKNZW18} &3.32&0.33& 0.25&0.63\\
DR(eye)VE \cite{DBLP:journals/pami/PalazziACSC19}&2.27&0.45& 0.32&0.64\\
ACLNet \cite{DBLP:journals/pami/WangSXCLB21} &2.51&0.35& 0.35&0.64\\
DADA \cite{DBLP:journals/tits/FangYQXY22} &2.19&\textbf{0.50}& \textbf{0.37}&0.66\\
DRIVE \cite{DBLP:conf/iccv/Bao0K21} &2.65&0.33& 0.19&0.66\\
\hline
CAP$_{\text{-Text+Att+1}}$ &2.02&0.39&0.28& 0.82\\
CAP$_{\text{-Text+Att+3}}$ &\textbf{1.88}&0.39&0.28& \textbf{0.85}\\
 \hline
  \end{tabular}}
  \label{tab6}
  \end{table}

\begin{figure}[!t]
  \centering
 \includegraphics[width=\linewidth]{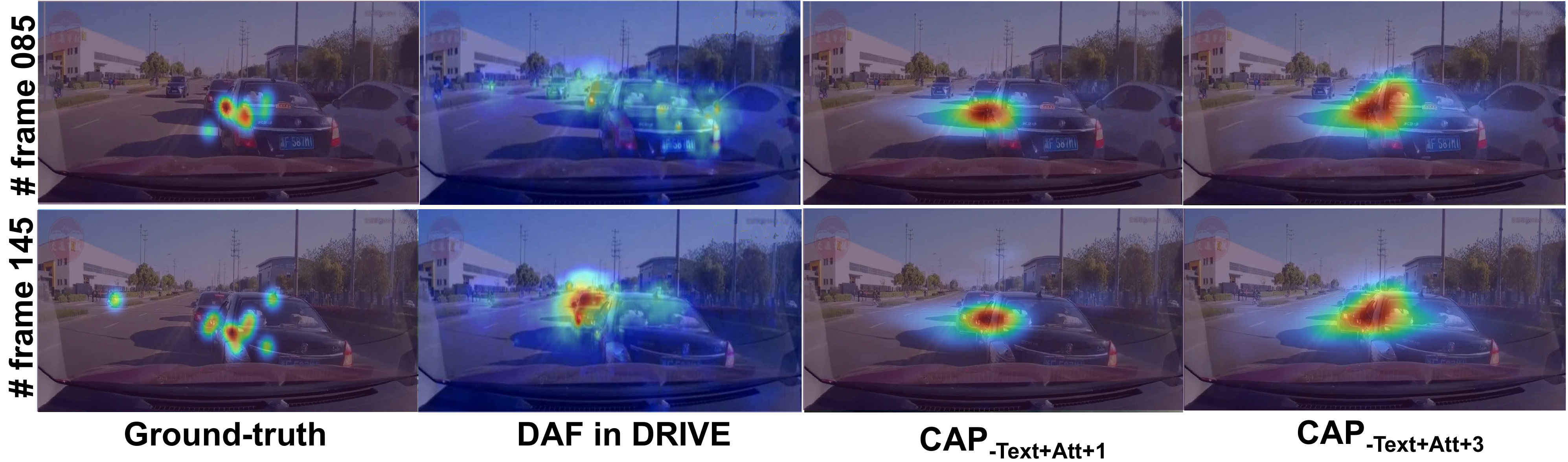}
  \caption{Two frames with driver attention maps obtained by CAP$_{\text{-Text-Att+1}}$, CAP$_{\text{-Text+Att+3}}$, and the dynamic attention map of DRIVE \cite{DBLP:conf/iccv/Bao0K21}.}
  \label{fig10}
\end{figure}

 We also check the driver attention reconstruction performance with different layers of T2I-SFLayers. The evaluation baselines are noted as CAP$_{\text{-Text-Att+1}}$ and CAP$_{\text{-Text+Att+3}}$. We also take many state-of-the-art methods, such as BDDA \cite{DBLP:conf/accv/XiaZKNZW18}, DR(eye)VE \cite{DBLP:journals/pami/PalazziACSC19}, ACLNet \cite{DBLP:journals/pami/WangSXCLB21} DADA \cite{DBLP:journals/tits/FangYQXY22}, and DRIVE \cite{DBLP:conf/iccv/Bao0K21} in this evaluation. Table. \ref{tab6} presents the results, which show that our method obtains the best KLdiv value and s-AUC value. Compared with the methods specially designed for driver attention reconstruction, our method seems not to be the best for the CC and SIM metrics, which means that the reconstructed driver attention maybe with an inconsistent distribution shape with the ground truth but the larger s-AUC value of our method has the better fixation localization than other methods. As for the related work DRIVE \cite{DBLP:conf/iccv/Bao0K21}, our method shows manifested performance improvement. In addition, we find that more layers of T2I-SFLayers benefit the performance. We also present some qualitative results in Fig. \ref{fig10}, where the attention maps with three T2I-SFLayers can cover more fixation points than one T2I-SFLayer version.

\subsection{Overall Accident Prediction Results}
We compare the proposed method with several state-of-the-art approaches in performance comparison. They are DSA-RNN \cite{DBLP:conf/accv/ChanCXS16}, AdaLEA \cite{DBLP:conf/cvpr/SuzukiKAS18}, UncertaintyTA \cite{bao2020uncertainty}, and DRIVE \cite{DBLP:conf/iccv/Bao0K21} with the available source code. The results on the MINI-Test of these methods are taken from DRIVE \cite{DBLP:conf/iccv/Bao0K21}, and we re-implement DRIVE on FULL-Test by training DRIVE on the FULL-Train set with 50 epochs, where the driver attention map of each frame is also obtained by MLNet \cite{cornia2016deep}.

\begin{table}[!t]\footnotesize
  \centering
  \caption{Performance comparison of accident prediction on MINI-Test and FULL-Test of CAP-DATA. The best is marked by \textbf{bold} font.}
\renewcommand{\arraystretch}{1.2}
     \setlength{\tabcolsep}{1.6mm}{
\begin{tabular}{cccccc}
\toprule[0.8pt]
      \multirow{2}[4]{*}{} &Baselines&AUC$\uparrow$  & AP$\uparrow$&TTA$_{0.5}\uparrow$& mTTA$\uparrow$\\
       \hline
     \multirow{2}[4]{*}{}&DSA-RNN \cite{DBLP:conf/accv/ChanCXS16}&0.47& -&3.095&-\\
  \multirow{2}[4]{*}{MINI}& AdaLEA  \cite{DBLP:conf/cvpr/SuzukiKAS18}&0.55& -&3.890&-\\
  \multirow{2}[4]{*}{}&UncertaintyTA \cite{bao2020uncertainty}&0.60&-& 3.849&-\\ 
  \multirow{2}[4]{*}{}&DRIVE\cite{DBLP:conf/iccv/Bao0K21}&0.69& 0.72&3.657&4.295\\ 
  \cline{2-6}
  \multirow{2}[4]{*}{}&CAP$_{\text{+Text+Att+3}}$&0.82&\textbf{0.75}& 3.951&4.464\\
  \multirow{2}[4]{*}{}&CAP$_{\text{-Text+Att+3}}$&\textbf{0.83}&0.74& \textbf{4.020}&\textbf{4.497}\\
\toprule[0.8pt]
    \multirow{2}[4]{*}{FULL-5s}&DRIVE\cite{DBLP:conf/iccv/Bao0K21}&0.70& 0.90&3.436&4.204\\
    \multirow{2}[4]{*}{}&CAP$_\text{{+Text+Att+3}}$ &\textbf{0.80}&\textbf{0.92}&3.612 & \textbf{4.303}\\
  \multirow{2}[4]{*}{}&CAP$_{\text{-Text+Att+3}}$&\textbf{0.80}&0.86& \textbf{3.673}& 4.102\\
    \hline
    \multirow{2}[4]{*}{FULL-4s}&DRIVE\cite{DBLP:conf/iccv/Bao0K21}&0.77& 0.90&2.480&3.290\\
    \multirow{2}[4]{*}{}&CAP$_\text{{+Text+Att+3}}$ &\textbf{0.84}&\textbf{0.93}& \textbf{2.818}& \textbf{3.535}\\
  \multirow{2}[4]{*}{}&CAP$_{\text{-Text+Att+3}}$&\textbf{0.84}&0.92& 2.777& 3.533\\
    \hline
    \multirow{2}[4]{*}{FULL-2s}&DRIVE\cite{DBLP:conf/iccv/Bao0K21}&0.67& 0.65&0.631&1.090\\
    \multirow{2}[4]{*}{}&CAP$_\text{{+Text+Att+3}}$ &\textbf{0.80}&0.65& \textbf{1.078}& \textbf{1.767}\\
  \multirow{2}[4]{*}{}&CAP$_{\text{-Text+Att+3}}$&0.77&\textbf{0.69}&1.069& 1.764\\
 \hline
  \end{tabular}}
  \label{tab7}
  \end{table}

 \begin{table*}[!t]\scriptsize
  \centering
  \caption{Performance evaluation on the main ego-car involved accidents in prediction, where the accident categories with top-5 amount, top-7 amount, and top-8 amount of categories are evaluated for future \textbf{5s-prediction}, \textbf{4s-prediction}, and \textbf{2s-prediction}, respectively. The categories are denoted in the footnote of this table.  ``\textbf{(category index)}/(A/B)" in the table denotes the accident category with A and B denoting the number of positive and negative samples.}
\begin{tabular}{cccc|ccc|ccc|ccc|ccc}
\toprule[0.8pt]
   \multirow{2}[4]{*}{Baselines} & \multicolumn{3}{c|}{\cellcolor{lightgray!40}5s-\textbf{(8)}/(70/0)} & \multicolumn{3}{c|}{\cellcolor{lightgray!30}5s-\textbf{(10)}/(110/10)} & \multicolumn{3}{c|}{\cellcolor{lightgray!30}5s-\textbf{(11)}/(686/6)}& \multicolumn{3}{c|}{\cellcolor{lightgray!30}5s-\textbf{(12)}/(38/0)}& \multicolumn{3}{c}{\cellcolor{lightgray!30}5s-\textbf{(14)}/(97/0)}\\
\cmidrule{2-16}          & AP & TTA$_{0.5}$& mTTA & AP& TTA$_{0.5}$&mTTA & AP & TTA$_{0.5}$&mTTA& AP & TTA$_{0.5}$&mTTA & AP & TTA$_{0.5}$&mTTA\\
\hline
DRIVE \cite{DBLP:conf/iccv/Bao0K21} &\textbf{0.99}&3.447& 4.084&0.91&3.515&4.3461& 0.97&3.416&4.228&0.97& 3.110 &4.272&0.96& 3.376 &4.362\\
\hline
CAP$_{\text{-Text}}$  &0.98&3.684& 4.459&0.89&3.697&\textbf{4.414}& 0.97&3.611&4.257&\textbf{1.00}& 3.606 &4.403&\textbf{1.00}& 3.376 &\textbf{4.372}\\
CAP$_{\text{+Text}}$  &\textbf{0.99}&\textbf{3.797}& \textbf{4.618}&\textbf{0.92}&\textbf{3.867}&4.215& \textbf{0.97}&\textbf{3.623}&\textbf{4.288}&0.99& \textbf{3.624} &\textbf{4.415}&0.97& \textbf{3.474} &4.324\\
\toprule[0.8pt]
   \multirow{2}[4]{*}{Baselines} & \multicolumn{3}{c|}{\cellcolor{lightgray}4s-\textbf{(6)}/(56/0)} & \multicolumn{3}{c|}{\cellcolor{lightgray}4s-\textbf{(8)}/(121/23)} & \multicolumn{3}{c|}{\cellcolor{lightgray}4s-\textbf{(10)}/(196/6)}& \multicolumn{3}{c|}{\cellcolor{lightgray}4s-\textbf{(11)}/(1108/109)}& \multicolumn{3}{c}{\cellcolor{lightgray}4s-\textbf{(12)}/(70/7)}\\
\cmidrule{2-16}          & AP & TTA$_{0.5}$& mTTA & AP& TTA$_{0.5}$&mTTA & AP & TTA$_{0.5}$&mTTA& AP & TTA$_{0.5}$&mTTA & AP & TTA$_{0.5}$&mTTA\\
\hline
DRIVE\cite{DBLP:conf/iccv/Bao0K21} &0.96&2.562& 3.314&0.81&2.448&3.289& 0.96&2.529&3.308&0.90& 2.474 &3.294&0.89& 2.606 &3.323\\
\hline
CAP$_{\text{-Text}}$  &0.96&2.977& 3.483&\textbf{0.86}&2.755&3.570& \textbf{0.97}&2.848&3.586&\textbf{0.93}& 2.772 &\textbf{3.519}&0.89& \textbf{2.899} &3.396\\
CAP$_{\text{+Text}}$  &\textbf{0.98}&\textbf{3.010}& \textbf{3.596}&0.85&\textbf{2.850}&\textbf{3.654}& \textbf{0.97}&\textbf{2.947}&\textbf{3.745}&\textbf{0.93}& \textbf{2.796} &3.406&\textbf{0.94}& 2.863 &\textbf{3.409}\\
\toprule[0.8pt]
   \multirow{2}[4]{*}{Baselines} & \multicolumn{3}{c|}{\cellcolor{lightgray}4s-\textbf{(13)}/(51/8)} & \multicolumn{3}{c|}{\cellcolor{lightgray}4s-\textbf{(14)}/(168/35)} & \multicolumn{3}{c|}{\cellcolor{gray!80}2s-\textbf{(5)}/(155/62)}& \multicolumn{3}{c|}{\cellcolor{gray!80}2s-\textbf{(6)}/(169/73)}& \multicolumn{3}{c}{\cellcolor{gray!80}2s-\textbf{(8)}/(295/108)}\\
\cmidrule{2-16}          & AP & TTA$_{0.5}$& mTTA & AP& TTA$_{0.5}$&mTTA & AP & TTA$_{0.5}$&mTTA& AP & TTA$_{0.5}$&mTTA & AP & TTA$_{0.5}$&mTTA\\
\hline
DRIVE\cite{DBLP:conf/iccv/Bao0K21} &0.83&2.471& 3.294&0.79&2.434&3.286& 0.53&0.684&1.126&0.52& 0.650 &1.102&0.64& 0.636 &1.091\\
\hline
CAP$_{\text{-Text}}$  &\textbf{0.87}&2.737& 3.579&\textbf{0.82}&2.655&3.370& 0.56&1.135&1.738&\textbf{0.54}& 1.088 &1.766&0.61& 1.073 &1.796\\
CAP$_{\text{+Text}}$  &0.84&\textbf{2.764}& \textbf{3.686}&0.79&\textbf{2.698}&\textbf{3.535}& \textbf{0.58}&\textbf{1.184}&\textbf{1.809}&\textbf{0.54}& \textbf{1.113} &\textbf{1.778}&\textbf{0.63}& \textbf{1.079} &\textbf{1.837}\\
\toprule[0.8pt]
   \multirow{2}[4]{*}{Baselines} & \multicolumn{3}{c|}{\cellcolor{gray!80}2s-\textbf{(10)}/(626/229)} & \multicolumn{3}{c|}{\cellcolor{gray!80}2s-\textbf{(11)}/(3145/1234)} & \multicolumn{3}{c|}{\cellcolor{gray!80}2s-\textbf{(12)}/(189/73)}& \multicolumn{3}{c|}{\cellcolor{gray!80}2s-\textbf{(13)}/(125/46)}& \multicolumn{3}{c}{\cellcolor{gray!80}2s-\textbf{(14)}/(331/142)}\\
\cmidrule{2-16}          & AP & TTA$_{0.5}$& mTTA & AP& TTA$_{0.5}$&mTTA & AP & TTA$_{0.5}$&mTTA& AP & TTA$_{0.5}$&mTTA & AP & TTA$_{0.5}$&mTTA\\
\hline
DRIVE\cite{DBLP:conf/iccv/Bao0K21} &0.62&0.636& 1.094&0.60&0.635&1.093& 0.51&0.621&1.085&0.63&0.635 &1.093&0.57& 0.656 &1.104\\
\hline
CAP$_{\text{-Text}}$  &0.59&1.061& \textbf{1.784}&\textbf{0.61}&1.068&\textbf{1.738}& 0.51&1.071&\textbf{1.758}&\textbf{0.66}& \textbf{1.075} &1.704&\textbf{0.58}& \textbf{1.078} &1.709\\
CAP$_{\text{+Text}}$  &\textbf{0.63}&\textbf{1.106}& 1.781&\textbf{0.61}&\textbf{1.077}&1.729& \textbf{0.55}&\textbf{1.079}&1.727&0.60& 1.039 &\textbf{1.754}&0.57& 1.072 &\textbf{1.723}\\
\bottomrule
  \end{tabular}
\begin{tablenotes}
\item \scriptsize{The number in the bold brackets in each column represent the ego-car involved accident category index. The categories here are \textbf{(5)} \emph{ego-car hits a crossing motorbike}, \textbf{(6)} \emph{ego-car hits a motorbike}, \textbf{(8)} \emph{ego-car hits a truck}, \textbf{(10)} \emph{ego-car hits a crossing car}, \textbf{(11)} \emph{ego-car hits a car}, \textbf{(12)} \emph{ego-car is overtaken by a car}, \textbf{(13)} \emph{ego-car hits large roadblocks}, and \textbf{(14)} \emph{ego-car hits a curb}.}
\end{tablenotes}
  \label{tab8}
  \end{table*}

\textbf{On MINI-Test:}
The evaluation of the MINI-Test is shown in Table. \ref{tab7}. It can be observed that our method achieve a significant performance improvement. The time length of all testing samples in MINI-Test is 5 seconds (5s), where the predicted TTA$_{0.5}$ exceeds 4 seconds (4s) in our method and mTTA achieves 4.497 seconds, which has a large gap over the best method DRIVE \cite{DBLP:conf/iccv/Bao0K21}. Because DSA-RNN \cite{DBLP:conf/accv/ChanCXS16}, AdaLEA \cite{DBLP:conf/cvpr/SuzukiKAS18}, and UncertaintyTA \cite{bao2020uncertainty} need to pre-detect the participants, where the detection error may influence the accident prediction, the AUC scores are not good as compared with DRIVE and our method. Notably, CAP$_{\text{-Text+Att+3}}$ is a little better than CAP$_{\text{+Text+Att+3}}$ here, which means the text information is leveraged to the vision information. Different from DRIVE which obtains the driver attention map offline, our method can obtain a significant improvement on the AUC value by jointly optimizing the driver attention map reconstruction and accident prediction.

\textbf{On FULL-Test:}
Based on the testing evaluation for FULL-Test, we check the performance on the cases of 5s, 4s, and 2s prediction of accidents. Here, we take DRIVE \cite{DBLP:conf/iccv/Bao0K21} as a baseline because it has the same consideration of the driver attention. From Table. \ref{tab7}, we can see that 5s and 4s predictions show promising AP values and exceed 0.85 for all cases, while 2s prediction has relatively lower AP value. The underlying reason is that owning to the video length restriction, we have about $\frac{1}{3}$ negative samples in all testing set for 2s prediction case, while 5s and 4s predictions only have about $\frac{1}{10}$ negative samples (see Table. \ref{tab3}). Therefore, the ratio of negative samples has an influence for the AP value. As for the AUC values, our method can obtain a relatively stable performance for all prediction cases. In particular, DRIVE seems to be vulnerable to the ratio of negative samples, where the AUC value degrades to 0.67 for the 2s prediction case. 

As for TTA metrics, our method still shows a better performance than that of DRIVE. With the increase of testing samples from 5s to 2s predictions, the TTA$_{0.5}$ gap between our method and DRIVE increases from $0.6$s to $0.85$s, and mTTA gap increases from about $0.2$s to $1.02$s. Hence, from this phenomenon, the ratio of negative samples and the number of testing samples have an impact on the TTA metrics and AP score. In addition, based on the Text-to-Image Shift Fusion layer (T2I-SFLayer), CAP$_{\text{-Text+Att+3}}$ shows comparable performance with CAP$_{\text{+Text+Att+3}}$.

\subsection{Evaluation on Different Attributes}
In this subsection, we analyze the performance on different attributes, where the influence of accident category, occasion adaptability, and road type specificity are analyzed. Notably, the versions of our method adopt three layers of T2I-SFLayer and the driver attention, while the subscript omits the ``Att+3" for a clear description in the following.

\subsubsection{Accident Category Influence}
In common sense, ego-car-involved accidents threaten safe driving and could be avoided with an early occurrence warning. Therefore, we analyze several typical types of ego-car-involved accidents here. Similarly, the cases of 5s, 4s, and 2s predictions are shown and the comparison results with DRIVE \cite{DBLP:conf/iccv/Bao0K21} are listed in Table. \ref{tab8}. Notably, because of the imbalance of negative and positive samples, some categories here have no negative samples. Hence, the AUC metric is not used in this evaluation. From the results in Table. \ref{tab8}, the phenomena of high AP value with few negative samples is clearly shown. For the categories without negative samples, the AP values are close to $1$ and degrade apparently with more ratio of negative samples. TTA metrics are computed on the positive samples, which are stable for different accident categories. Therefore, for ego-car involved accident categories, the performance is stable with about $\pm0.1$ difference for TTA metrics. 

From the prediction cases, we find that categories of \emph{(10) ``ego-car hits a crossing car"} and \emph{ (5)``ego-car hits a crossing motorbike"} obtain the relative best TTA$_{0.5}$ value in all categories. Therefore, the keyword \emph{``crossing"} could be treated as a signal for accident prediction. Similarly, the text information is successfully shifted to the vision path and comparable performance is obtained for CAP$_{\text{-Text}}$ and CAP$_{\text{+Text}}$. Compared with the dynamic road participants, the accident categories of \emph{(13) ``ego-car hits large roadblock"} and \emph{ (14)``ego-car hits a curb"} involve static obstacles. TTA$_{0.5}$ results of these two categories generally are smaller than other ones. These two findings indicate that the \textbf{interaction} between ego-car with road participants is important for accident prediction. For example, because the static roadblocks only have a distance interaction with the ego-car, the collision warning is determined by the velocity of the ego-car and the distance to the roadblock. On the contrary, dynamic road participants have a behavior interaction, which may provide the more useful signal for accident prediction.

 \begin{figure}[!t]
  \centering
 \includegraphics[width=\linewidth]{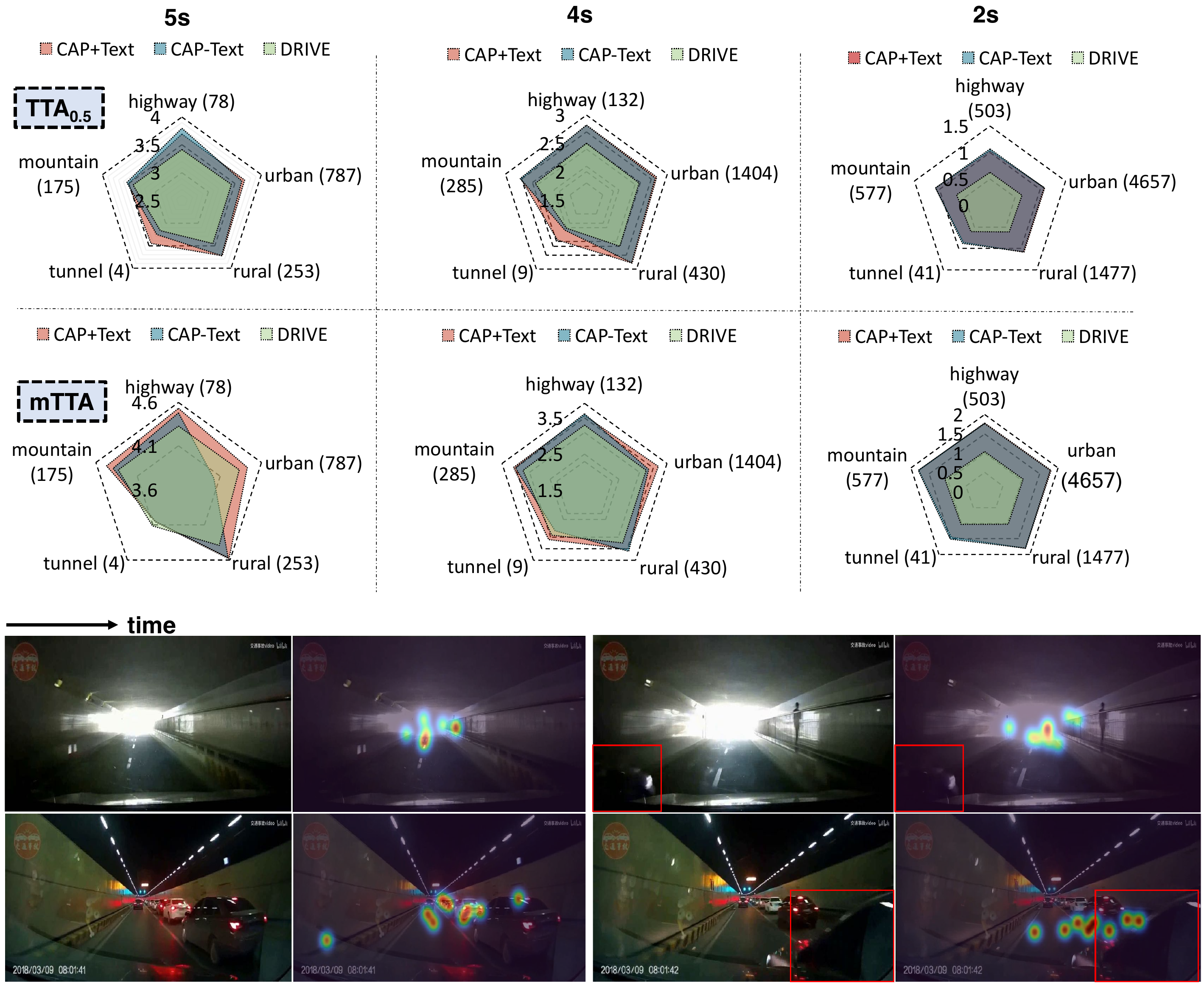}
  \caption{The radar graphs of TTA$_{0.5}$ and mTTA scores on the cases of 5s, 4s, 2s accident predictions with respect to the occasion adaptability for CAP$_{\text{-Text}}$, CAP$_{\text{+Text}}$, and DRIVE \cite{DBLP:conf/iccv/Bao0K21}. There are also two tunnel scenarios overlapped the driver attention map, where the crashing cars are marked by \textcolor{red}{red} boxes.}
  \label{fig11}
\end{figure}

\subsubsection{Occasion Adaptability}
Different road occasions have differing driving conditions. As for driving activity, the safety level is also different. We evaluate the TTA metrics (TTA$_{0.5}$ and mTTA) on highway, urban, rural, mountain, and tunnel occasions for our method and DRIVE \cite{DBLP:conf/iccv/Bao0K21}, where TTA metrics only need the positive samples. The performance is shown by the radar graphs in Fig. \ref{fig11}. From the results, we can see that our method is superior to DRIVE to a large extent from the cases of 5s prediction to 2s prediction. As a common result, all comparison approaches show a relatively weak performance on the tunnel occasion compared with other occasions. Although the number of samples in the tunnel scene takes the smallest ratio, this phenomenon is common. To analyze the reason, we present two tunnel samples in Fig. \ref{fig11}, where the light condition in the tunnel scene has a rather large impact on the driver attention. Taking the first tunnel scene in Fig. \ref{fig11} as an example, near the tunnel exit, the large intensity contrast of the white light makes the driver blinded to find the overtaking car. This phenomenon is called ``tunnel vision" with a steep decline in visual performance, which is investigated by some previous studies \cite{ringer2016impairing,yang2022beyond}. On the contrary, the second tunnel scene in Fig. \ref{fig11} has a better light condition, where the overtaking car is noticed by the driver fixations. Consequently, to reduce traffic accident, road facilities (\eg, good lighting conditions in tunnel scenes) are also important. 

\subsubsection{Road Type Specificity}
\begin{figure}[!t]
  \centering
 \includegraphics[width=\linewidth]{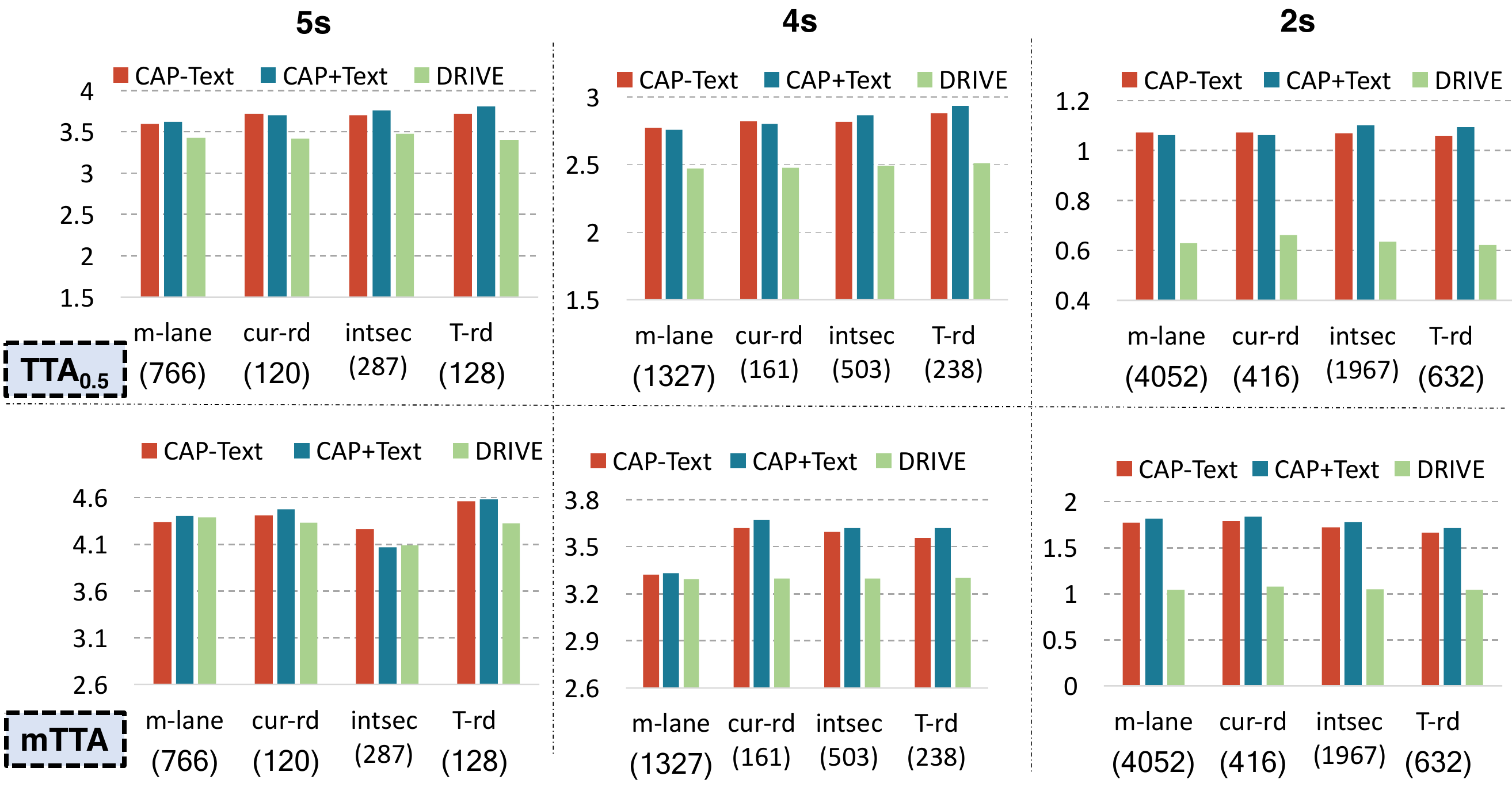}
  \caption{The histograms of TTA$_{0.5}$ and mTTA values on the cases of 5s, 4s, 2s accident predictions on the road type specificity for CAP$_{\text{-Text}}$, CAP$_{\text{+Text}}$, and DRIVE \cite{DBLP:conf/iccv/Bao0K21}.}
  \label{fig12}
\end{figure}

We also evaluate accident prediction performance on road type specificity of main-lane (m-lane), curve road (cur-rd), intersection (intsec), and T-road (T-rd). Similarly, the TTA metrics on DRIVE and our methods with (CAP$_{+\text{Text}}$) and without (CAP$_{-\text{Text}}$) the text input are compared. Fig. \ref{fig12} presents the histogram results and demonstrates that the performance gap between our method and DRIVE increases from the cases of 5s prediction to 2s prediction. Besides, the accident prediction ability of different methods seems not to be sensitive for different road types from the TTA$_{0.5}$ scores. As for mTTA values, our method shows relatively better performance on the cur-rd and T-rd for the long-term (5s or 4s) accident prediction. Compared with the main lane and intersection, curve-rd and T-rd have manifested direction guidance, and the driver will focus on the turning vanishing point of the road. On the contrary, the short-term (2s) accident prediction shows little performance difference because there is no time for considering other things.
\subsection{Evaluation on the CCD Dataset}

In this work, we also evaluate the performance of the CCD Dataset. The pre-defined testing set in the CCD Dataset has 900 positive samples without the negative sample. Each sample has $50$ video frames for 5 seconds of duration. In this evaluation, DSA-RNN \cite{DBLP:conf/accv/ChanCXS16}, UncertaintyTA \cite{bao2020uncertainty}, and DSTA \cite{DBLP:journals/tits/KarimLQY22}, and XAI-ANT \cite{DBLP:journals/corr/abs-2108-00273} are compared and trained with 3600 videos. Notably, we implement our method and DRIVE \cite{DBLP:conf/iccv/Bao0K21} for a cross-dataset evaluation, which is trained on FULL-Train set (with only 1,020 raw videos annotated driver attention) of our CAP-DATA. Table. \ref{tab9} presents the prediction results. Because there is no negative sample in CCD testing evaluation, AP scores of all methods are very high, which is also verified by the aforementioned analysis. In addition, 
our method shows comparable performance with other approaches even with a cross-dataset evaluation.
\vspace{-1.5em}
 \begin{table}[!t]\small
  \centering
  \caption{Performance comparison of accident prediction on the testing set in CCD dataset. The best results are marked by \textbf{bold} font.}
\begin{tabular}{ccc}
\toprule[0.8pt]
Baselines&AP$\uparrow$& mTTA$\uparrow$\\
       \hline
DSA-RNN \cite{DBLP:conf/accv/ChanCXS16}&0.996& 4.52\\
UncertaintyTA \cite{bao2020uncertainty}&0.995&4.74\\ 
DSTA \cite{DBLP:journals/tits/KarimLQY22}&0.996&\textbf{4.87}\\ 
XAI-ANT \cite{DBLP:journals/corr/abs-2108-00273}&0.940&4.57\\
\hline
DRIVE $_{(\text{DADA-2000}\rightarrow \text{CCD})}$\cite{DBLP:conf/iccv/Bao0K21}&0.992& 4.49\\ 
${\text{CAP}_{\text{-Text+Att+3}}}_{(\text{DADA-2000}\rightarrow \text{CCD})}$&\textbf{0.997}&4.63\\
 \hline
  \end{tabular}
  \label{tab9}
  \end{table}
\section{Conclusions}
\label{con}
This work proposes a Cognitive Accident Prediction (CAP) method with the consideration of driver attention and text descriptions. CAP is also a vision-text fusion model which explores the text-to-vision shift for adapting to the practical testing phase without text description. 
A large-scale benchmark with 11,727 videos of 2.19 million frames are constructed with the annotation of the accident time window, fact-effort-reason-introspection descriptions (\ie, the test description before accident, accident category description, reason description, and prevention advice), and many static attributes, \eg, weather, road type, and occasions. 
Based on this benchmark, CAP is exhaustively evaluated by comparison with other state-of-the-art approaches, and larger Time-to-Accident (TTA) and accuracy are obtained. In comparison, we find that AP metric and mTTA may not be good evaluation metrics, where AP is vulnerable to the ratio of negative samples and mTTA seems to be not discriminating for different driving situations. In addition, after checking the performance with or without text input, we find that the \emph{attentive text-to-vision shift fusion module} works and comparable performance is obtained by testing the trained model without text input. In the future, we will further explore the interaction between the road participants and the ego-vehicle, the novel view, \ie, drone \cite{zheng2020university}, road navigation map, and the counterfactual analysis for the ``cause-effort" problem in accident prediction. 

{\small
\bibliographystyle{IEEEtran}
\bibliography{ref}
}

\end{document}